\documentclass{article}
\usepackage{microtype}
\usepackage{graphicx}
\usepackage{multirow}
\usepackage{subcaption}
\usepackage{booktabs}
\usepackage[usenames, dvipsnames, table]{xcolor}
\usepackage{svg}
\setsvg{inkscape=inkscape}
\usepackage{enumitem}
\usepackage{subcaption}
\usepackage{booktabs} 
\usepackage[pagebackref=true]{hyperref}
\renewcommand*{\backref}[1]{}
\renewcommand*{\backrefalt}[4]{{%
    \ifcase #1 Not cited.%
          \or Cited on page~#2.%
          \else Cited on pages #2.%
    \fi%
    }}

\usepackage[most]{tcolorbox}
\usepackage{mdframed}
\colorlet{LightGray}{White!98!Periwinkle}

\usepackage{thmtools}

\newtcolorbox{hypothesis}{%
  enhanced jigsaw,
  colback=LightGray,
  drop shadow,
  boxrule=0.9pt,
  boxsep=0.1pt,
  left=4pt,
  right=4pt,
  top=4pt,
  bottom=4pt
}

\usepackage{hyperref}

\newcommand{\method}[0]{\textsc{SimpleMix}}

\usepackage[accepted]{icml2025}

\usepackage{amsmath}
\usepackage{amssymb}
\usepackage{mathtools}
\usepackage{amsthm}
\usepackage{comment}

\usepackage[capitalize,noabbrev]{cleveref}

\theoremstyle{plain}

\theoremstyle{definition}

\theoremstyle{remark}

\usepackage[textsize=tiny]{todonotes}

\icmltitlerunning{\method: Frustratingly Simple Off- and On-policy Data Mixing in Language Model Preference Learning}

\begin{document}

\twocolumn[
\icmltitle{\method: Frustratingly Simple Mixing of Off- and On-policy Data \\ in Language Model Preference Learning}



\icmlsetsymbol{equal}{*}

\begin{icmlauthorlist}
\icmlauthor{Tianjian Li}{jhu}
\icmlauthor{Daniel Khashabi}{jhu}
\end{icmlauthorlist}

\icmlaffiliation{jhu}{Center for Language and Speech Processing, Johns Hopkins University, Baltimore, US}

\icmlcorrespondingauthor{Tianjian Li}{tli104@jhu.edu}

\icmlkeywords{Machine Learning, ICML}

\vskip 0.3in
]



\printAffiliationsAndNotice{\icmlEqualContribution} 

\begin{abstract}
Aligning language models with human preferences relies on pairwise preference datasets. While some studies suggest that on-policy data consistently outperforms \emph{off}-policy data for preference learning, others indicate that the advantages of \emph{on}-policy data may be task-dependent, highlighting the need for a systematic exploration of their interplay.

In this work, we show that on-policy and off-policy data offer \emph{complementary} strengths in preference optimization: on-policy data is particularly effective for reasoning tasks like math and coding, while off-policy data performs better on open-ended tasks such as creative writing and making personal recommendations. Guided by these findings, we introduce \method, an approach to combine the complementary strengths of on-policy and off-policy preference learning by simply mixing these two data sources. Our empirical results across diverse tasks and benchmarks demonstrate that \method{} substantially improves language model alignment. Specifically, \method{} improves upon on-policy DPO and off-policy DPO by an average of 6.03\% on Alpaca Eval 2.0. Moreover, it outperforms prior approaches that are much more complex in combining on- and off-policy data, such as HyPO and DPO-Mix-P, by an average of 3.05\%. 


\end{abstract}

\section{Introduction}
\label{introduction}



Alignment of Language Models (LMs) has bestowed them with the ability to learn better from demonstrations \cite{brown2020language}, extrapolate from reasoning chains \cite{wei2022chain}, and produce responses aligned with human values \cite{ouyang2022training}. 


The ongoing debate \cite{ivison2024unpacking, tajwar2024preference, tang2024understandingperformancegaponline, pmlr-v235-xu24h} in the literature that compares alignment with on-policy data (data is sampled from the LM to be aligned) with off-policy data (data is sampled not from the LM to be aligned). The debate arrives at mismatched conclusions, with some works reporting on-policy outperforms off-policy \cite{tajwar2024preference, pmlr-v235-xu24h} while others reporting that the gains from on-policy training are minimal, and sometimes even underperform off-policy training \cite{ivison2024unpacking, ahmadian-etal-2024-back, lambert2024tulu3pushingfrontiers}. Existing work \cite{ivison2024unpacking, pmlr-v235-xu24h} that compares on- and off-policy data does not control using the same algorithm under different data sources, and does not investigate the impact of task types \cite{tang2024understandingperformancegaponline}. 

To address these weaknesses, we vary \emph{only} the data source and break down the performances by different task types to answer the following research questions:


\newcommand{\subscript}[2]{$#1 _ #2$}

\begin{enumerate}[noitemsep, label=(\subscript{Q}{{\arabic*}}),leftmargin=*]
    \item \label{q1} \textit{Under what circumstances on- vs off-policy data offer different strengths?}
    \item \label{q2} \textit{Can we leverage their complementarity for more efficient alignment?}
\end{enumerate}

\begin{figure*}
    \centering
    \includegraphics[width=\textwidth]{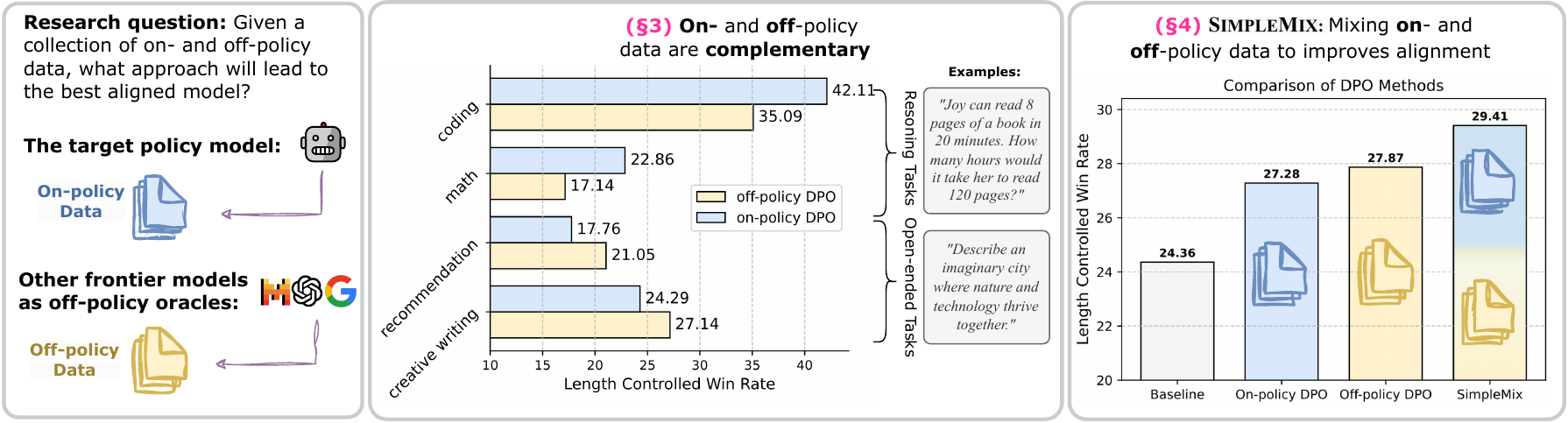}
    \caption{\textbf{Left panel:} Our work studies data origin in preference optimization of LMs. \textbf{Middle panel:} We show that on-policy data and off-policy data are complementary: on-policy data mostly improves the model's performance on reasoning tasks that are objectively correct or incorrect (e.g., Math and Coding) while off-policy data improves on sub-tasks where humans might disagree with each other (e.g., creative writing and personal recommendation) (\S \ref{Section:3}). \textbf{Right panel:} Our proposed method \method~mixes on and off-policy data, outperforming solely using either on or off-policy data. (\S \ref{Section:4})}
    \label{fig:teaser}
\end{figure*}


With regards to \ref{q1}, we observe that on-policy data and off-policy data offer complementary strengths in preference learning of LMs: \textit{On-policy} data is most effective in tasks that tend to have objective answers and require reasoning skills, e.g., math and coding. \textit{Off-policy} data is effective in more open-ended tasks, e.g., creative writing and making personal recommendations, where it is common to have multiple differing preferred responses (\S\ref{Section:3}).

To address \ref{q2}, we propose \method: to mix on- and off-policy data for preference learning. Our approach consistently improves upon only using on- or off-policy data solely in data-constrained setups where the same amount of responses is used for each method. Furthermore, we show that \method~is able to outperform prior approaches that mix on- and off-policy data in more complex ways \cite{song2024hypo, shi2024crucialrolesamplersonline}, showcasing the effectiveness of our method despite its simplicity (\S\ref{Section:4}).

Figure \ref{fig:teaser} illustrates our work: The complementary nature of on- and off-policy data in LM preference learning: on-policy DPO \cite{rafailov2024direct} outperforms off-policy DPO in reasoning tasks, e.g., solving a math problem, while underperforms in open-ended tasks, e.g., describing an imaginary city(\S \ref{Section:3}); \method: Mixing on- and off-policy data in DPO can outperform DPO with a single data source (\S \ref{Section:4}). \method{} achieves an average improvement of 6.03 over on-policy and off-policy methods on Alpaca Eval 2.0 and outperforms existing hybrid approaches by 3.05 while maintaining simplicity.

To sum up, our contribution is two-fold:

(1) We show that on- and off-policy data in preference learning are complementary: on-policy data excels at tasks requiring reasoning and objective verification, such as math and coding, whereas off-policy data is more effective for open-ended tasks, such as creative writing and personal recommendations.

(2) We demonstrate that a good balance between on- and off-policy data consistently outperforms approaches relying solely on either data source.

\section{Preliminaries and Notations}

We consider post-training of LMs. Given a user prompt $x$ in natural language,  the language model $\pi(\cdot|x)$ takes the prompt as input and outputs a probability distribution over natural language responses $y$. The Supervised Fine-Tuning (SFT) stage of LM updates a pre-trained LM $\pi_\text{base}$ on a dataset with (prompt, response) pairs: $\mathcal{D}_\textrm{SFT} = \{(x_1, y_1), \hdots, (x_N, y_N)\}$ with the following maximum likelihood objective: $\pi_\textrm{SFT} \in \arg \max_{\pi} \prod_{i=1}^N \pi(y_i \mid x_i)$. 

\paragraph{Preference Optimization (PO)}

While SFT is a step forward in aligning LMs to follow human instructions, the resulting model is not aligned to human preferences.
For example, humans often prefer structured, concise yet complete responses. 
To align $\pi_\text{SFT}$ to human preferences, we perform Preference Optimization (PO) on top of $\pi_\text{SFT}$ with a pairwise preference distribution: $\mathcal{D} = \{(x, y_w, y_l) ... \}$, where there are two responses $(y_w, y_l)$ for a single prompt. $y_w$ is preferred by human annotators (usually referred to as the ``winning" or ``chosen" response), and $y_l$ is dispreferred (usually referred to as the ``losing" or ``rejected" response). In essence, PO aims to steer $\pi_\textbf{SFT}$ towards the winning responses that are preferred by humans. Formally, the objective is defined as \cite{rafailov2024direct}:
\begin{align}
\mathbb{E}_{(x, y_w, y_l) \sim \mathcal{D}} \Bigg[ 
\log \sigma\Bigg( \beta &\log \frac{\pi_\theta(y_w \mid x)}{\pi_\text{SFT}(y_w \mid x)} \nonumber \\
&- \beta \log \frac{\pi_{\theta}(y_l \mid x)}{\pi_\textrm{SFT}(y_l \mid x)} \Bigg) 
\Bigg],
\label{eq:DPO}
\end{align}
where $\sigma$ is a non-decreasing function. Maximizing (1) should maximize the increase in log-likelihood of $y_w$ and the decrease of $y_l$. Ideally, the model learns to generalize characteristics that make $y_w$ preferred. For example, if $y_w$ are constructed that are factually correct and $y_l$ contains hallucinations, the hope is that by steering the policy towards a few factually correct responses, the policy generalizes so that it assigns high probability to \textit{all} responses that are factually correct. We differentiate between on- and off-policy PO depending on $\mathcal{D}$: from which $y_w$ and $y_l$ are sampled.

\paragraph{On-policy PO} Assuming that the LM $\pi$ is parameterized by $\theta$. If the responses $y$ are sampled from a policy that is derived from either $\pi_\text{SFT}$ or $\pi_{\theta}$, we describe $y$ as \textbf{on-policy data}: where ${\mathcal{D} = \mathcal{D}_\text{on} = \{(x, y) \mid y\sim\pi_\theta \text{~or~} \pi_\text{SFT}\}}$.\footnote{Our definition of ``on-policy" coincides with \citet{lambert2024tulu3pushingfrontiers}. Note that minor differences exist between our definition of on- vs. off-policy and the traditional Reinforcement Learning literature, which treats responses (actions) that are not sampled from $\pi_\theta$ as off-policy \cite{song2024hypo, xie2024exploratorypreferenceoptimizationharnessing, xiong2024iterative}.} Notable examples of on-policy PO include Proximal Policy Optimization (PPO) \cite{schulman2017proximal} that optimizes a mathematically equivalent objective as equation (1) using on-policy generations $y \sim \pi_{\theta}$, and ``on-policy DPO" \cite{guo2024directlanguagemodelalignment} that uses a stronger language model as a judge to label pairs of on-policy generations and optimizes equation (1) with $y_w, y_l \sim \pi_{\theta}$. 

\paragraph{Off-policy PO} Recall that on-policy data are sampled from a policy that is derived from either $\pi_\text{SFT}$ or $\pi_{\theta}$, we describe data that are not on-policy data as off-policy data and PO performed on top of an off-policy dataset ${\mathcal{D} = \mathcal{D}_\text{off}}$ Usually, off-policy data are sampled from a collection of open-sourced or proprietary models \citep{pmlr-v235-cui24f, wang2024helpsteer2, bai2022training}. A notable example of off-policy PO is Direct Preference Optimization \citep{rafailov2024direct} (DPO) which directly optimizes equation (1) with the responses sampled from an off-policy dataset. 
\vspace{-10pt}


\section{On- vs. Off-policy Data are Complementary}
\label{Section:3}

In this section, we vary the data source in DPO, whether $\mathcal{D} = \mathcal{D}_\text{on}$ or $\mathcal{D} = \mathcal{D}_\text{off}$ in equation \ref{eq:DPO} to answer \ref{q1}. Our setup ensures a consistent setup by only varying the data source while \textit{fixing} the alignment algorithm, in contrast to prior work that does not control the alignment algorithm by comparing on- vs. off-policy data.

\paragraph{Our Hypothesis} We found that in \citet{ivison2024unpacking}, PPO shows most gains on GSM8k \cite{cobbe2021gsm8k}, a dataset consisting of grade-school math questions, while being similar in performance to DPO on general tasks such as MMLU \cite{hendrycks2020measuring}. In \citet{tang2024understandingperformancegaponline}, the performance gap between on- and off-policy data is smallest on chatbot arena side by side \cite{chiang2024chatbotarenaopenplatform}, which also attests to the policy's performance on general human queries. We thus make the following hypothesis:

\begin{hypothesis}
\textbf{Hypothesis:} On-policy data mostly helps in tasks that are objectively correct or not (e.g., math), whereas the difference between on- and off-policy data on open-ended tasks (e.g., creative writing) is minimal.
\end{hypothesis}

We describe our experiment setup in \S \ref{section:3.1} and report our results that validate our hypothesis in \S \ref{section:3.2}.




\begin{figure*}[t]
\centering
    \begin{subfigure}[b]{0.44\textwidth}
    \centering
    \includegraphics[scale=0.45]{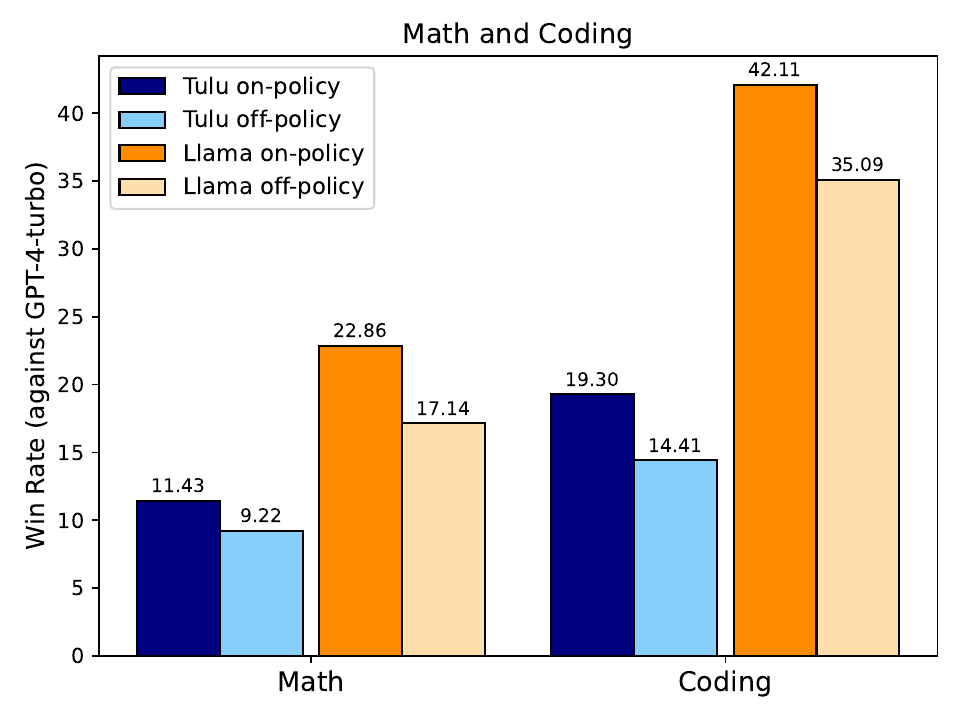}
    \label{fig:llama3_per_category}
    \end{subfigure}
    \hspace{0.02\textwidth}
    \begin{subfigure}[b]{0.44\textwidth}
    \centering
    \includegraphics[scale=0.45]{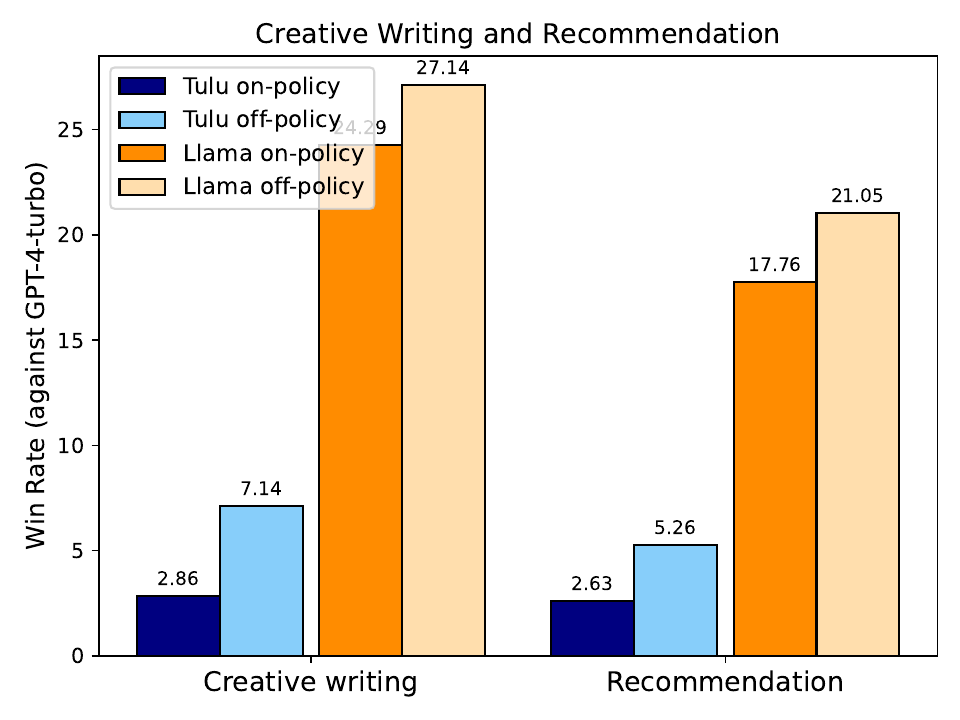}
    \label{fig:tulu_per_category}
    \end{subfigure}
    \caption{Comparison of win rates (against GPT-4-turbo) across different prompt categories in Alpaca Eval 2.0 for (left) objective tasks that have a groundtruth answer and (right) open-ended tasks where humans have individual preferences. \textbf{On-policy DPO improves performance in math and coding, while off-policy DPO demonstrates better performance in creative writing and making personal recommendations.}}
    \label{fig:winrates_by_category}
\end{figure*}

\begin{figure*}[h!]
\centering
   \hspace{-0.5cm}\begin{subfigure}[b]{0.44\textwidth}
    \centering
    \includegraphics[scale=0.48]{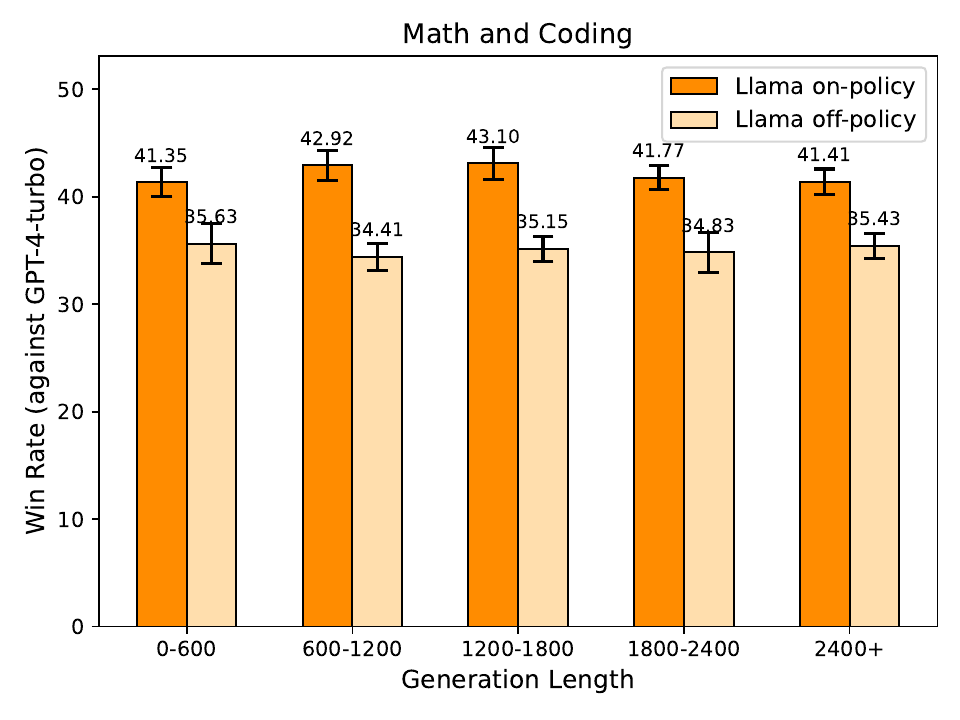}
    \label{fig:llama3_per_category_sm}
    \end{subfigure}
    \hspace{0.02\textwidth}
    \begin{subfigure}[b]{0.44\textwidth}
    \centering
    \includegraphics[scale=0.48]{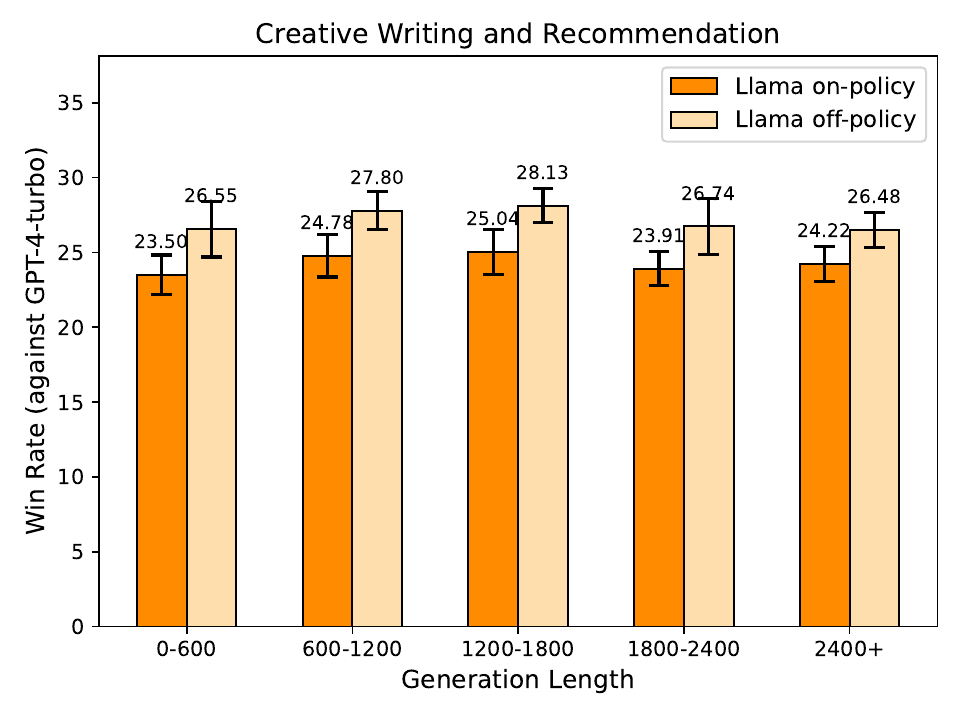}
    \label{fig:tulu_per_category_sm}
    \end{subfigure}
    \vspace{-1cm}
    \caption{Comparison of win rates (against GPT-4-turbo) by the length of generation. On-policy training does not significantly outperform off-policy training as generation length increases in both math and coding tasks (left) and creative writing as well as recommendation tasks (right). 
    Error bars show 95\% confidence intervals from bootstrapping.
    }
    \label{fig:winrate_by_length}
\end{figure*}


\subsection{Experiment Setup}
\label{section:3.1}
\paragraph{Models and Dataset} We use two language models as $\pi_\text{SFT}$: \texttt{meta-llama-3.1-8B-Instruct} \citep{dubey2024llama3herdmodels} and \texttt{Llama-3.1-Tulu-3-8B-SFT} \citep{lambert2024tulu3pushingfrontiers}. We experiment on the UltraFeedback dataset \cite{pmlr-v235-cui24f} that consists of 60k prompts. For each prompt, two responses were collected from a model pool. We treat the two responses as \textbf{off-policy} data. For each prompt, we sample $N$ responses\footnote{Sampling Parameters: temperature $\tau=0.7$, top-p $p=0.9$} as \textbf{on-policy} data. We use the best reward model \texttt{Skywork/Skywork-Reward-Gemma-2-27B-v0.2} in RewardBench \cite{lambert2024rewardbench}\footnote{Best performing as of 2024-11-27} as a proxy of human judgment to score the pair of responses. We refer to this model as the \emph{oracle} reward model. The response with the highest reward is labeled as $y_w$ and the response with the lowest reward is labeled as $y_l$. More detailed hyperparameters are reported in Appendix~\ref{appendix:hparams}.

\paragraph{Evaluation} To test our hypothesis, we break down the prompts in Alpaca Eval 2.0 \citep{alpaca_eval} into different categories. A detailed breakdown of prompt category and distribution can be found in Appendix~\ref{appendix:alpaca_eval}. We select four representative tasks: two are objective tasks that can be verified: \textbf{mathematical reasoning or calculation (Math)} and \textbf{Coding}, and the other two tasks are tasks that require modeling human open-ended preference: \textbf{Creative writing} and \textbf{making personal recommendations (Recommendation)}. We report the length controlled win rate \cite{dubois2024lengthcontrolled} of each selected task for models trained with on- and off-policy DPO.

\subsection{Results}
\label{section:3.2}
\paragraph{When does on-policy sampling help?} 

Figure \ref{fig:winrates_by_category} shows the win rate for the 4 representative categories. We plot the detailed performance of the performance of on- and off-policy DPO on Math and Coding prompts in Alpaca Eval 2.0 at left of Figure \ref{fig:winrates_by_category}, and we plot the results in Creative writing and Recommendation queries at Figure \ref{fig:winrates_by_category}.

We observe that \textit{on-policy DPO helps in objective and verifiable tasks.} Specifically, in our experiments of \texttt{meta-llama-3.1-8B-Instruct}, on-policy DPO outperforms off-policy DPO in math and coding, improving the length-controlled win rate by \textcolor{Green}{+5.72\%} and \textcolor{Green}{+7.02\%} respectively. However, on open-ended tasks: creative writing and making personal recommendations, on-policy DPO underperforms off-policy DPO by \textcolor{red}{-2.85\%} and \textcolor{red}{-3.29\%} respectively. 

\paragraph{Is the improvement about length?}

One might argue that a possible explanation is that tasks requiring longer generations benefit more from on-policy data and that the improvement in math and coding can be attributed to the improvement in length. To investigate this, we plot the histogram of the length controlled win rate by the length of the generated sequences in Figure \ref{fig:winrate_by_length}. 

We observe that on-policy DPO does not outperform off-policy DPO as the generation length increases. This indicates that the improvements in on-policy training on math and coding cannot be solely attributed to increasing the length of the generations. The type of task is the primary factor in determining whether on-policy or off-policy training is more effective.

We conclude that on-policy DPO mostly improves on objective tasks (e.g., math and coding) while off-policy DPO excels in improving open-ended tasks (e.g. creative writing and personal recommendation) and that this distinction cannot be explained by generation length. In the next section, we demonstrate how combining the strengths of on-policy data with the efficiency and abundance of off-policy data available on the web can lead to improved LM alignment.

\section{\method: Simply Mixing On- and Off-policy Data Improves Alignment}
\label{Section:4}

\begin{table*}
\centering
\begin{tabular}{lll}
\toprule
\textbf{Method Name} & \textbf{Sampler of $y_w, y_l$} & \textbf{Loss Function} \\
\midrule
 \rowcolor[gray]{0.9}     \multicolumn{3}{c}{ \textit{Non-Hybrid Methods}}  \\ \midrule
Off-policy DPO & $y_w, y_l \sim \mathcal{D} $ & DPO($y_w, y_l$) \\
On-policy DPO & $y_w, y_l \sim \pi_\text{SFT}$ & DPO($y_w, y_l$) \\
\midrule
 \rowcolor[gray]{0.9}     \multicolumn{3}{c}{ \textit{Hybrid Methods}}  \\ \midrule
DPO-Mix-P \citep{shi2024crucialrolesamplersonline} & $y_w, y_l \sim \{\pi_\theta^{3/2}\pi_\text{SFT}^{-1/2}, \pi_\theta^{1/2}\pi_\text{SFT}^{1/2}\}$ & DPO($y_w, y_l$) \\
HyPO \citep{song2024hypo} & $y_w, y_l \sim \mathcal{D}, y \sim \pi_\theta$ & DPO($y_w, y_l$) - $\lambda \log(y \mid x)\cdot \text{sg}^*\left(\frac{\pi_{\theta}(y \mid x)}{\pi_\text{SFT}(y \mid x)}\right)$ \\
\rowcolor{cyan!20} \method~(Ours) & $y_w, y_l \sim \{\mathcal{D}, \pi_\text{SFT}\}$ & DPO($y_w, y_l$) \\ 
\bottomrule
\end{tabular}
\caption{A comparison between \method~and other baselines we compare in this work. $^* \text{sg}()$ denotes the stop gradient operation.}
\label{tab:baselines}
\end{table*}

In this section, we show the benefits of combining on- and off-policy data in preference optimization. Specifically, we employ the data source $\mathcal{D} = \mathcal{D}_\text{on} \cup \mathcal{D}_\text{off}$ in equation \ref{eq:DPO} to answer \ref{q2}. We describe our experiment setup and baselines at \S \ref{subsection:4.1}, and report our main results at \S \ref{subsection:4.2}.

\subsection{Setup}
\label{subsection:4.1}


\paragraph{\method} Our method \method~combines on- and off-policy data, where the winning $y_w$ and losing responses $y_l$ are sampled from $\pi_\text{SFT}$ and the off-policy dataset $\mathcal{D}$ with \textbf{equal} probability. An in-depth study on the effect of different sampling probabilities is at \S \ref{subsection:5.1}.

\paragraph{Baselines} We compare the following baselines with \method. A more detailed comparison of our baselines with \method is in Table \ref{tab:baselines}.

\begin{itemize}[leftmargin=*]
    \item \textbf{On-policy DPO}: Winning and losing response sampled from the SFT model: $y_w, y_l \sim \pi_\text{SFT}(\cdot \mid x)$.
    \item \textbf{Off-policy DPO}: Winning and losing response sampled from an off-policy dataset: $y_w, y_l \sim \mathcal{D}_\text{off}$.
    \item \textbf{HyPO} \citep{song2024hypo}: Off-policy DPO with KL Regularization using on-policy data: $y_w, y_l \sim \mathcal{D}_\text{off}, y \sim \pi_\theta(\cdot \mid x)$. Compared to HyPO \cite{song2024hypo}, \method~changes the online KL regularization objective to a DPO loss. 
    \item \textbf{DPO-Mix-P} \citep{shi2024crucialrolesamplersonline}: On-policy data generated with a interpolation between the current model $\pi_\theta$ and $\pi_\textrm{SFT}$.
    $y_w, y_l \sim \{\pi_\theta^{3/2}\pi_\text{SFT}^{-1/2}, \pi_\theta^{1/2}\pi_\text{SFT}^{1/2}\}$. Compared to DPO-Mix-P \cite{shi2024crucialrolesamplersonline}, \method~can be seen as a ``hard"  version of sampling from an interpolation between an off-policy LM and the SFT policy $\pi_\text{SFT}$ with equal weights.
\end{itemize}

\paragraph{Models and Datasets} We experiment with two $\pi_\textrm{SFT}$ models \texttt{meta-llama/Llama-3.1-8B-Instruct} and \texttt{allenai/Llama-3.1-Tulu-3-8B-SFT} with DPO \citep{rafailov2024direct} and experiment on two pairwise preference datasets: Ultrafeedback \cite{pmlr-v235-cui24f} and HelpSteer2 \cite{wang2024helpsteer2}. A detailed description of the collection of prompts and responses in both datasets can be found in Appendix~\ref{appendix:dataset_details}. We train our models for one epoch on the same amount of $(y_w, y_l)$ pairs for every baseline. This controls that every method has seen the same amount of data, although on-policy DPO requires more computing since it requires sampling from either $\pi_\text{SFT}$ or $\pi_{\theta}$. 

\paragraph{Evaluation} To evaluate the aligned policy, we follow existing literature on LM alignment and use the same 7 benchmarks in the FineWeb \cite{penedo2024finewebdatasetsdecantingweb} evaluation suite, containing Commonsense QA (CQA; \citet{talmor-etal-2019-commonsenseqa}), Hellaswag (HS; \citet{zellers2019hellaswag}), Openbook QA (OQA; \citet{OpenBookQA2018}), PIQA \citep{bisk2020piqa}, Winograde \citep{sakaguchi2020winogrande}, ARC-Challenge (ARC-C; \citet{clark2018think}), and MMLU \citep{hendrycks2020measuring}, which covers 7 knowledge and common sense based benchmarks. We also use Ifeval \citep{zhou2023instruction} and Alpaca Eval 2.0 \citep{dubois2023alpacafarm, dubois2024lengthcontrolled} for evaluating general instruction following ability. We report the average scores of the seven tasks + Ifeval prompt level loose accuracy as \textbf{Avg. Score} and Alpaca Eval 2.0 length controlled win rate (LC). Detailed descriptions of benchmarks can be found in Appendix \ref{appendix:eval_details}.

\subsection{Main Results}
\label{subsection:4.2}

Table \ref{tab:hybrid} shows the experiment results on Alpaca Eval 2.0 and averaged scores over 8 benchmarks (Avg. Score) for all of our baselines. We report the detailed breakdown of individual benchmarks in Appendix~\ref{appendix:full_results}. 

We find that \textbf{\method~consistently yields the best performance across both models and evaluation metrics.} This indicates that the widely available source of off-policy preference data can still provide valuable signals to preference optimization. We plot the per-task performance in Alpaca Eval 2.0 of models trained with \method~in Figure \ref{fig:winrates_by_category_with_sm} in Appendix \ref{appendix:sm_per_task}. \method~offers a balance between the performance on reasoning (math and coding) and open-ended tasks (creative writing and recommendation).


\begin{table*}
\centering
\begin{tabular}{@{}lcccc@{}}
\toprule
 & \multicolumn{2}{c}{$\pi_\text{SFT}=$ \textbf{Tulu-3-8B-SFT}} & \multicolumn{2}{c}{$\pi_\text{SFT}=$ \textbf{Llama-3.1-8B-Instruct}} \\ 
\cmidrule(l){2-5}
 & Alpaca Eval 2.0 LC (\%) $\uparrow$ & Avg. Score $\uparrow$ 
 & Alpaca Eval 2.0 LC (\%) $\uparrow$ & Avg. Score $\uparrow$ \\
\midrule
$\pi_{\text{SFT}}$   & 8.52 & 59.72 & 24.36 & 60.88 \\ \midrule
 \rowcolor[gray]{0.9}     \multicolumn{5}{c}{ \textit{Experiments on UltraFeedback \cite{pmlr-v235-cui24f}}}  \\ \midrule
Off-policy DPO       & 14.23 & 60.02 & 27.87 & 61.67 \\
On-policy DPO        & 15.17 & 59.70 & 27.28 & 62.09 \\ \midrule 
HyPO \cite{song2024hypo}                & 18.11 & 60.03 & 27.91 & 62.01 \\
DPO-Mix-P \cite{shi2024crucialrolesamplersonline}           & 17.25 & 60.70 & 27.79 & 61.61 \\ 

\rowcolor{cyan!20} \method              & \textbf{20.64} & \textbf{61.14} & \textbf{29.41} & \textbf{63.06} \\  \rowcolor[gray]{0.9}   \midrule   \multicolumn{5}{c}{ \textit{Experiments on HelpSteer2 \cite{wang2024helpsteer2}}} \\ \midrule
Off-policy DPO       & 14.74  & 59.94 & 24.81 & 61.54 \\
On-policy DPO        & 15.26 & 60.21 & 25.09 &  61.77 \\ \midrule 
HyPO \cite{song2024hypo}                & 18.19 & 59.96 & 27.15 & 61.69 \\
DPO-Mix-P \cite{shi2024crucialrolesamplersonline}           & 17.75 & 60.04 & 27.11  & 61.49 \\ 

\rowcolor{cyan!20} \method           & \textbf{21.11} & \textbf{61.22} & \textbf{29.12} & \textbf{62.12}
\\ 
\bottomrule
\end{tabular}
\caption{Performance comparison among various on and off-policy preference optimization methods on UltraFeedback \cite{pmlr-v235-cui24f} and HelpSteer2 \cite{wang2024helpsteer2}. \textbf{Avg. Score} refers to the average score across 8 benchmarks listed in Appendix \ref{appendix:eval_details}. We found that the performance gap between on- and off-policy DPO is minimal. Incorporating off-policy data into on-policy DPO outperforms using on- or off-policy data only. Furthermore, \textbf{curating a balanced mixture of on- and off-policy data in DPO consistently yields the strongest performance in both LM-as-a-judge evaluation (Alpaca Eval 2.0) and reference-based evaluation (Avg. Score)}.}
\label{tab:hybrid}
\end{table*}

\section{Ablations in Preference Data Curation}

In this section, we investigate curation strategies in preference data curation. Specifically, we look into the impact of response diversity (\ref{subsection:5.1}), the mixture ratio between on- and off-policy data (\ref{subsection:5.2}), and the effect of filtering off-policy data (\ref{subsection:5.3}).

\subsection{The Effect of Preference Data Diversity}
\label{subsection:5.1}
Some works theoretically show that on-policy preference optimization benefits from more exploration \cite{rashidinejad2024sailheadwindalignmentrobust, song2024hypo, anonymous2024sample, xiong2024iterative}. This is to say that eliciting more diverse responses would be beneficial. A possible explanation of the effect of off-policy data is that it increases the diversity of generations. Works have reported that on-policy generations usually share the same prefix \citep{amr2.0-https://doi.org/10.35111/s444-np87} --- leading to reduced diversity among responses. On the other hand, off-policy data, esp. Ultrafeedback, the responses are collected from a \textbf{collection} of models rather than a single model, thus improving diversity.

To validate this claim, we compare our on+off policy mixture with directly increasing the diversity of responses with:

1. \textbf{Prompting:} We iteratively prompt the language model to generate $N$ responses for diversity \cite{lu2024benchmarkinglanguagemodelcreativity}: we first prompt the language model with $x$ and obtain the first response $y_1 \sim \pi(\cdot \mid x)$, then we condition on a written prompt $z$ that asks for a diverse response for the same input $y_2 \sim \pi(\cdot \mid x, z)$. We repeat this process to obtain $N$ responses. We report the prompt $z$ in Appendix \ref{appendix:diverse_prompt}. 
    
2. \textbf{Large temperature:} Sampling $y_w, y_l$ with a larger sampling temperature $\tau$. We experiment with $\tau = \{1, 2, 3\}$.

We sample $N=4$ responses from $\pi_\text{SFT} = \texttt{Llama-3.1-8B-Instruct}$ on the 60k prompts in Ultrafeedback, again we use the oracle reward model to annotate the best and worst response as $y_w$ and $y_l$. We report the results of training $\pi_\text{SFT}$ on this diverse response in Table \ref{tab:diverse_responses}. 

\begin{table}
\centering
\begin{tabular}{@{}lcc|c@{}}
\toprule
                  & LC (\%) $\uparrow$ & Avg. Score $\uparrow$ & Avg. Reward \\ \midrule
Prompting & 21.41                  & 58.27 & -10.75     \\ 
$\tau = 1.0$  &  26.94 & 60.21 & -9.44\\
$\tau = 2.0$  & 26.77 & 59.44 & -10.56\\
$\tau = 3.0$  & 26.18 & 58.83 & -10.77 \\
\method & \textbf{29.41}                  & \textbf{63.06} & \textbf{-3.39}     \\ \bottomrule
\end{tabular}
\caption{Performance comparison between different sampling methods for eliciting diverse responses. Explicitly prompting for diversity or increasing the sampling temperature $\tau$ results in generations with lower quality, as measured by our reward model. \textbf{Low-quality generations, albeit diverse, lead to worse performance on our evaluation when performing DPO on them.}}
\label{tab:diverse_responses}
\end{table}

\textbf{We found that eliciting more diverse responses often comes at the cost of decreasing generation quality, thus hurting performance.} We manually inspected the generations of prompting and larger generation temperature $\tau$, and we found that most of the generation, albeit diverse, often falls short in quality compared to direct sampling. The explanation is that diversity itself is not helpful because the diversity might fall into low-reward regions, increasing diversity while reducing quality. Our findings echo \cite{anonymous2025turning} who found that sampling with a larger temperature leads to degraded performance on various question-answering benchmarks.

\begin{figure*}[h!]
\centering
    \begin{subfigure}[b]{0.45\textwidth}
    \centering
    \includegraphics[width=\textwidth]{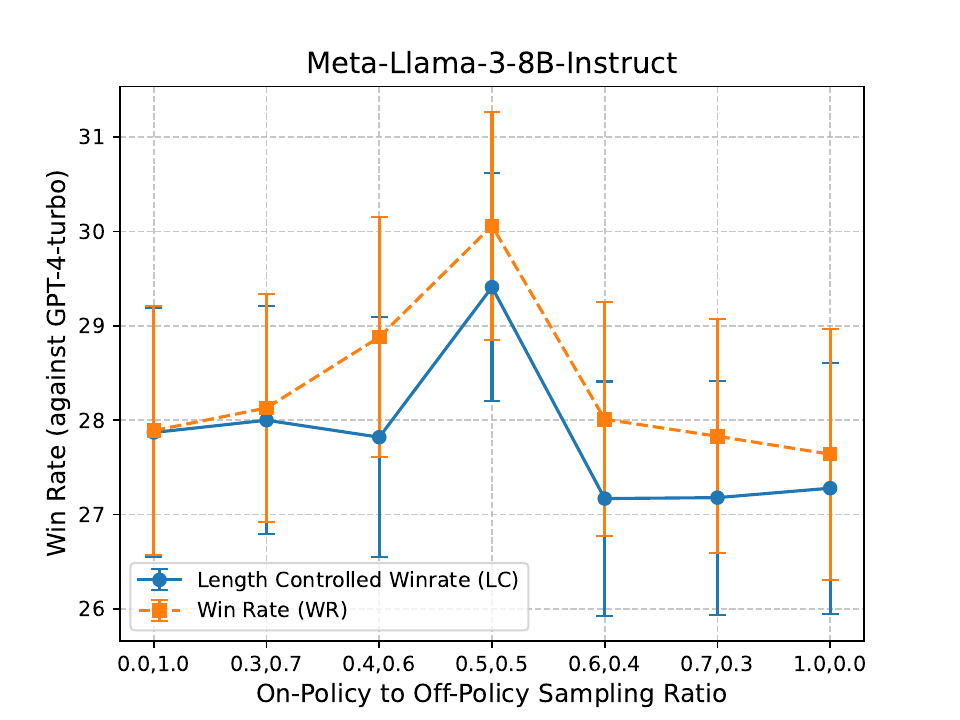}
    \label{fig:llama3_ratios}
    \end{subfigure}
    \hspace{0.02\textwidth}
    \begin{subfigure}[b]{0.45\textwidth}
    \centering
    \includegraphics[width=\textwidth]{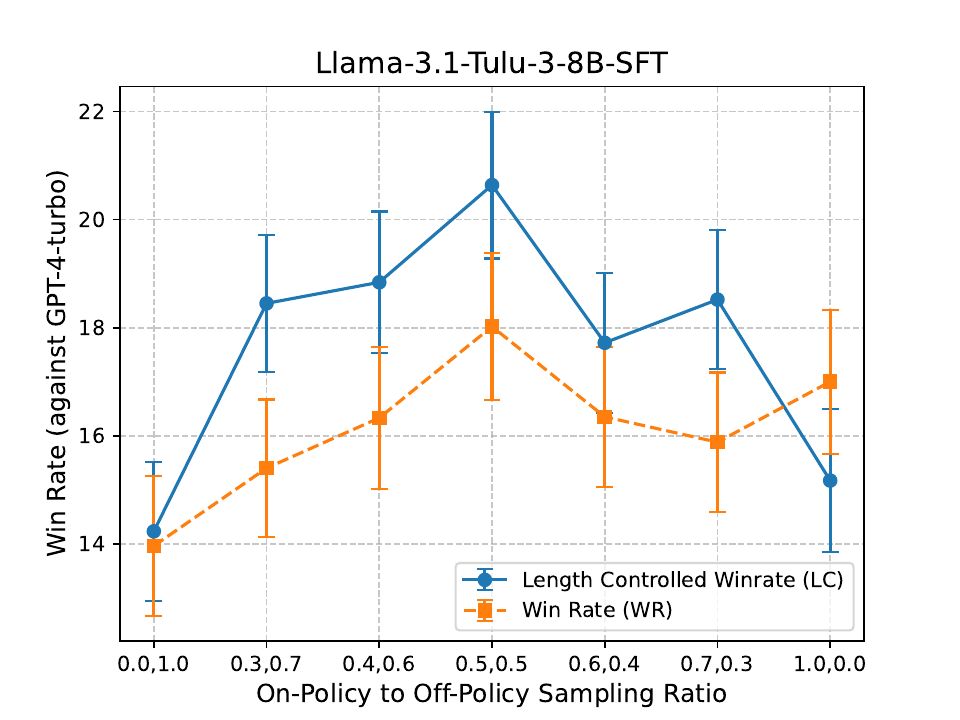}
    \label{fig:tulu_ratios}
    \end{subfigure}
   \caption{Perfomance on Alpaca Eval 2.0 for different on- to off-policy data ratio for performing DPO on top of $\pi_\text{SFT} = \texttt{Meta-LLama-3.1-8B-Instruct} \text{ (left) and } \texttt{Llama-3.1-Tulu-8B-SFT}$ (right) on the Ultrafeedback \cite{pmlr-v235-cui24f} dataset. \textbf{A balanced mixture (0.5 on-policy + 0.5 off-policy) outperforms other mixtures.}}
   \label{fig:wr_ratios}
\end{figure*}

\label{section:5}

\subsection{The Effect of On- vs. Off-policy Data Mixtures}
\label{subsection:5.2}
We further investigate different data mixtures of on- and off-policy data by keeping the total amount of data the same but varying the sampling ratio between on- and off-policy data. Figure \ref{fig:wr_ratios} shows the results of training \texttt{meta-Llama-3.1-8B-Instruct} and \texttt{Llama-3.1-Tulu-8B-SFT} on Ultrafeedback with different on- to off-policy data ratio. We observe that sampling with equal probability from both data sources (0.5 on-policy + 0.5 off-policy) outperforms other mixtures, which echos the results in \citet{pmlr-v202-ball23a} who finds that a mixture of 0.5 off-policy data + 0.5 on-policy data generally performs the best for traditional RL tasks. \citet{pmlr-v202-ball23a} also finds that in traditional RL tasks, a balanced 0.5 - 0.5 mixture is also the most stable. Unfortunately, based on the error bars in \ref{fig:wr_ratios}, we found no clear pattern about the variance of different mixtures.

\subsection{The Effect of Off-policy Data Filtering}
\label{subsection:5.3}

Off-policy responses are often collected with \textbf{various} open-sourced models for maximizing diversity, which often includes smaller models that are less capable. For example, responses in Ultrafeedback \cite{pmlr-v235-cui24f} are sampled from models ranging from the most capable GPT-4 to smaller open-sourced 7B models, often including low-quality responses. We investigate the effect of filtering out these low-quality responses when combined with on-policy data:

Specifically, we select a fraction $p$ of the entire Ultrafeedback dataset \cite{pmlr-v235-cui24f} according to the following heuristics:
\begin{itemize}[leftmargin=*]
    \item \textbf{Quality}: We annotate the pairs of generation with \texttt{Skywork-Reward-Gemma-2-27B-v0.2}. We select the top-$p$ pairs with the highest total reward.
    \item \textbf{On-policiness}: Existing work shows that LM learns better from on-policy data \cite{tajwar2024preference, tang2024understandingperformancegaponline}. We select top-$p$ of the most on-policy examples. We measure ``on-policiness" by the sum of the log-probabilities of chosen and rejected responses using $\pi_\text{SFT}$. 
    \item \textbf{Contrastiveness}: Existing work \cite{kim2024systematicexaminationpreferencelearning} reports that the policy learns better from highly contrastive pairs, where the quality gap between the chosen and rejected response is large. We select top-$p$ pairs where the difference between the chosen reward and the rejected reward is largest. 
    \item \textbf{Similarity}: Existing work \citep{pal2024smaugfixingfailuremodes, razin2024unintentional} points out that similar examples harms DPO performance. We take the cosine similarity of the last layer sentence embedding between the chosen and rejected response as a proxy for ``similarity" and select top-$p$ examples that are most dissimilar.
\end{itemize}

\begin{figure}[h!]
    \vspace{-5pt}
    \centering
    \includegraphics[scale=0.53]{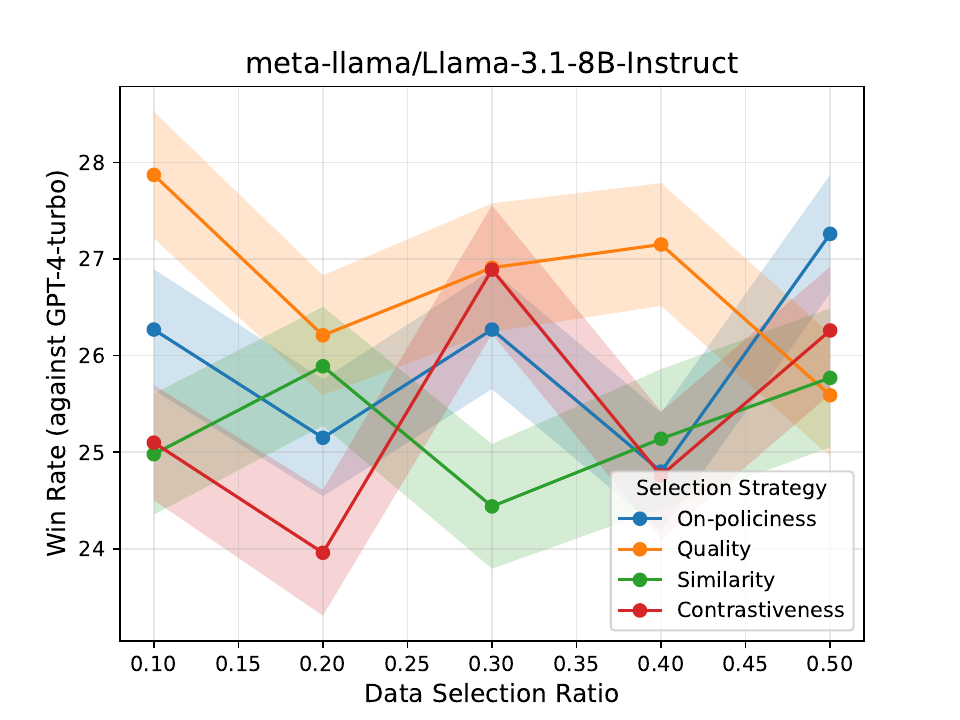}
    \caption{Alpaca Eval 2.0 length controlled win rate (LC) for different data selection strategies. \textbf{No criterion other than selecting high-quality data (measured by a reward model) significantly outperforms others for off-policy preference learning.}}
    \vspace{-5pt}
    \label{fig:criterion}
\end{figure}

We train \texttt{Llama-3.1-8B-Instruct} on filtered Ultrafeedback with $p = \{0.1, 0.2, 0.3, 0.4, 0.5\}$ and report the results on Alpaca Eval 2.0 in Figure \ref{fig:criterion}. We found that selecting data based on quality (\textcolor{orange}{orange}) outperforms other criteria except when $p = \{0.5\}$.

\begin{figure}
    \centering
    \hspace{-1pt}\includegraphics[scale=0.48]{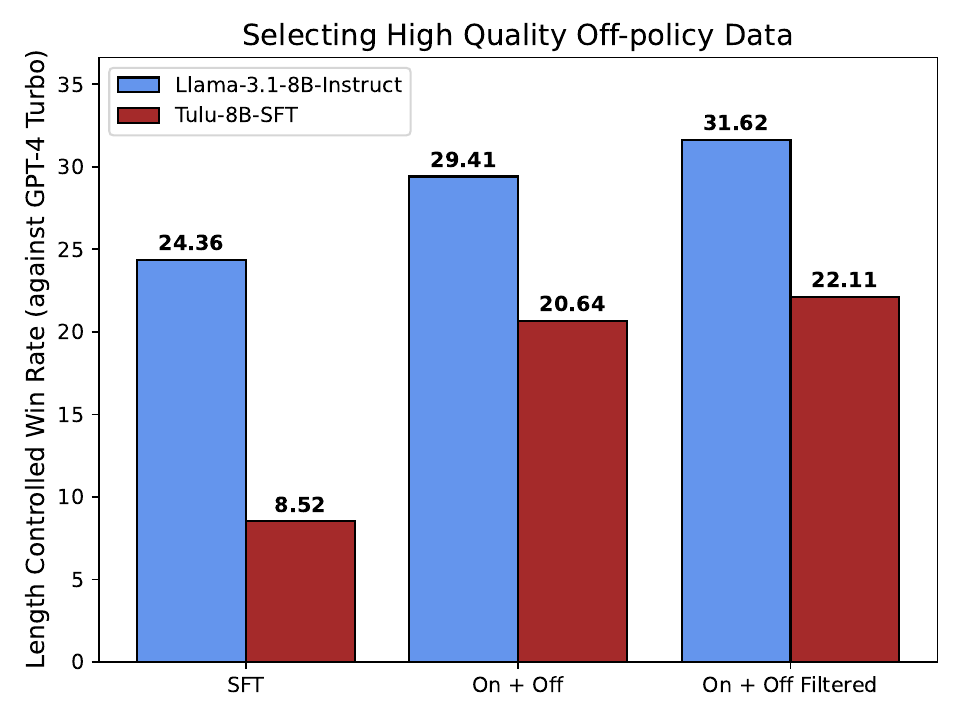}
    \caption{Alpaca Eval 2.0 Length Controlled Win Rate for $\pi_\text{SFT}$, Mixing On + Off-policy Data and Mixing in high-quality off-policy data with on-policy data in DPO. \textbf{Selecting high-quality off-policy data can improve performance using \method.}}
    \label{fig:filtered_dpo}
\end{figure}

Since we know that the quality of off-policy data has the most impact on performance, we filter the off-policy data according to its reward evaluated by our oracle reward model and mix it with the on-policy data. Specifically, we keep $p = 0.4$ fraction of Ultrafeedback and mix it with the same amount of on-policy data. We report the results in Figure \ref{fig:filtered_dpo}. We found that only keeping the high-quality off-policy data, when paired with on-policy data, can further improve the performance.

\section{Related Works}

\paragraph{Comparing On- and Off-Policy Alignment} 
The literature \cite{pmlr-v235-xu24h, ivison2024unpacking} usually compares PPO \cite{schulman2017proximal} with DPO \cite{rafailov2024direct}, arriving at mismatching conclusions with some works \cite{pmlr-v235-xu24h} showing that PPO outperforms DPO, some works show the gap between PPO and DPO is minimal \cite{lambert2024tulu3pushingfrontiers, ivison2024unpacking}. Two concurrent efforts are close to our work \cite{tajwar2024preference, tang2024understandingperformancegaponline}. Both studies investigate the differences between on and off-policy alignment, but \citet{tajwar2024preference} conducted experiments on a controlled setting that is different from real-world LM alignment, and although \citet{tang2024understandingperformancegaponline} have arrived at a conclusion that on-policy outperforms off-policy, the performance gap between on- and off-policy data in the most general task in their setting (chat arena side by side) is the smallest, which motivates us to investigate deeply. Additional related works on LM alignment is at Appendix \ref{appendix:additional_related_works}.
\paragraph{Hybrid RL}
While there has been many works that bridges on- and off-policy RL \cite{song2023hybrid, 10.5555/3618408.3619878, 10.5555/3294996.3295141, nakamoto2023calql, pmlr-v202-ball23a, lee2021offlinetoonline, tan2024hybrid, pmlr-v235-song24a}, few work studies combining off-policy and on-policy data in language model alignment \cite{song2024hypo, shi2024crucialrolesamplersonline, anonymous2024sera, bose2024hybridpreferenceoptimizationalignment}, whose contributions are mostly theoretical in terms of coverage \cite{song2024hypo}, convergence rates \cite{shi2024crucialrolesamplersonline, bose2024hybridpreferenceoptimizationalignment}, and optimality \cite{xiong2024iterative}. However, these works often only conduct experiments on LLM-as-a-judge benchmarks, which can be easily hacked \cite{singhal2024a, wei2024emojiattackmethodmisleading, park-etal-2024-offsetbias}, on models that are smaller and weaker (e.g., the Pythia suite). To the best of our knowledge, our work conducts the most comprehensive evaluation, including reference-based and LM-as-a-judge benchmarks, on top of state-of-the-art open-sourced models: llama-3.1-8B-Instruct \cite{dubey2024llama3herdmodels} and Tulu-3-8B-SFT \cite{lambert2024tulu3pushingfrontiers}.

\section{Limitations and Implications}

\paragraph{Limiations} We acknowledge that although we exhausted our resources to perform hyperparameter searching, the combinatorically large hyperparameter space makes it challenging to draw decisive conclusions. Therefore, although we observed that the performance gap between on- and off-policy preference optimization is minimal, it could be the case the we did not tune our model with the perfect configurations. Moreover, although we also exhausted our resources for evaluation on both reference-based and reference-free benchmarks, LLM-as-a-judge benchmarks can be hacked \cite{anonymous2025cheating}, exhibit certain biases \cite{panickssery2024llm, park-etal-2024-offsetbias}, and is sensitive to prompt formatting \cite{zheng2023judging}. Therefore, the applicability of our conclusions is limited to the hyperparameters configurations, models, training datasets and evaluation benchmarks on which we experimented.

\paragraph{Implications} Our work implies that data selection for LM alignment should be task dependent: for harder tasks where few high-quality data is available but the answer can be easily verifiable, e.g., math, where you can perform string matching for the answer., using on-policy data, outperforms off-policy data. However, in general, for tasks where abundant data is available online, performing on-policy sampling does not outperform. Therefore, one can resort to off-policy data, saving on the cost of generating multiple responses per query. We leave the impact of fine-grained ``on-policyness" of data on LM performance for future work.

\section{Conclusion}

In this paper, we show that in preference learning of LMs, \textit{on-policy} data and \textit{off-policy} data are complementary: on-policy data improves upon reasoning tasks that have a ground-truth answer, whereas off-policy data mostly improves upon more general tasks. We propose \method~: mixing up on- and off-policy data consistently improves performance when using data from a single source.

\section*{Impact Statement}
This paper presents work whose goal is to advance the field of 
Machine Learning. There are many potential societal consequences 
of our work, none of which we feel must be specifically highlighted here.

\section{Acknowledgments}
This work is supported by ONR grant (N00014-24-1-2089). 
We sincerely thank Jingyu Zhang, Weiting Tan, Marc Marone, Jeffery Cheng, and Orion Weller for fruitful discussions. We also thank the anonymous reviewers for their helpful suggestions.

\bibliography{icml2024/ref}

\providecommand{\CNFX}[1]{{\em{\textrm{(#1)}}}}
\begin{thebibliography}{85}
\providecommand{\natexlab}[1]{#1}
\providecommand{\url}[1]{\texttt{#1}}
\expandafter\ifx\csname urlstyle\endcsname\relax
  \providecommand{\doi}[1]{doi: #1}\else
  \providecommand{\doi}{doi: \begingroup \urlstyle{rm}\Url}\fi

\bibitem[Ahmadian et~al.(2024)Ahmadian, Cremer, Gall{\'e}, Fadaee, Kreutzer, Pietquin, {\"U}st{\"u}n, and Hooker]{ahmadian-etal-2024-back}
Ahmadian, A., Cremer, C., Gall{\'e}, M., Fadaee, M., Kreutzer, J., Pietquin, O., {\"U}st{\"u}n, A., and Hooker, S.
\newblock Back to basics: Revisiting {REINFORCE}-style optimization for learning from human feedback in {LLM}s.
\newblock In Ku, L.-W., Martins, A., and Srikumar, V. (eds.), \emph{Proceedings of the 62nd Annual Meeting of the Association for Computational Linguistics (Volume 1: Long Papers)}, pp.\  12248--12267, Bangkok, Thailand, August 2024. Association for Computational Linguistics.
\newblock \doi{10.18653/v1/2024.acl-long.662}.
\newblock URL \url{https://aclanthology.org/2024.acl-long.662}.

\bibitem[Almazrouei et~al.(2023)Almazrouei, Alobeidli, Alshamsi, Cappelli, Cojocaru, Debbah, Goffinet, Heslow, Launay, Malartic, Noune, Pannier, and Penedo]{falcon40b}
Almazrouei, E., Alobeidli, H., Alshamsi, A., Cappelli, A., Cojocaru, R., Debbah, M., Goffinet, E., Heslow, D., Launay, J., Malartic, Q., Noune, B., Pannier, B., and Penedo, G.
\newblock {Falcon-40B}: an open large language model with state-of-the-art performance.
\newblock 2023.

\bibitem[Anonymous(2024{\natexlab{a}})]{anonymous2024sample}
Anonymous.
\newblock Sample efficient alignment for {LLM}s.
\newblock In \emph{Submitted to The Thirteenth International Conference on Learning Representations}, 2024{\natexlab{a}}.
\newblock URL \url{https://openreview.net/forum?id=Pf8i7cv2CH}.
\newblock under review.

\bibitem[Anonymous(2024{\natexlab{b}})]{anonymous2024sera}
Anonymous.
\newblock Se{RA}: Self-reviewing and alignment of {LLM}s using implicit reward margins.
\newblock In \emph{Submitted to The Thirteenth International Conference on Learning Representations}, 2024{\natexlab{b}}.
\newblock URL \url{https://openreview.net/forum?id=uIGnuyDSB9}.
\newblock under review.

\bibitem[Anonymous(2025{\natexlab{a}})]{anonymous2025cheating}
Anonymous.
\newblock Cheating automatic {LLM} benchmarks: Null models achieve high win rates.
\newblock In \emph{The Thirteenth International Conference on Learning Representations}, 2025{\natexlab{a}}.
\newblock URL \url{https://openreview.net/forum?id=syThiTmWWm}.

\bibitem[Anonymous(2025{\natexlab{b}})]{anonymous2025turning}
Anonymous.
\newblock Turning up the heat: Min-p sampling for creative and coherent {LLM} outputs.
\newblock In \emph{The Thirteenth International Conference on Learning Representations}, 2025{\natexlab{b}}.
\newblock URL \url{https://openreview.net/forum?id=FBkpCyujtS}.

\bibitem[Azar et~al.(2024)Azar, Daniel~Guo, Piot, Munos, Rowland, Valko, and Calandriello]{pmlr-v238-gheshlaghi-azar24a}
Azar, M.~G., Daniel~Guo, Z., Piot, B., Munos, R., Rowland, M., Valko, M., and Calandriello, D.
\newblock A general theoretical paradigm to understand learning from human preferences.
\newblock In Dasgupta, S., Mandt, S., and Li, Y. (eds.), \emph{Proceedings of The 27th International Conference on Artificial Intelligence and Statistics}, volume 238 of \emph{Proceedings of Machine Learning Research}, pp.\  4447--4455. PMLR, 02--04 May 2024.
\newblock URL \url{https://proceedings.mlr.press/v238/gheshlaghi-azar24a.html}.

\bibitem[Bai et~al.(2022{\natexlab{a}})Bai, Jones, Ndousse, Askell, Chen, DasSarma, Drain, Fort, Ganguli, Henighan, et~al.]{bai2022training}
Bai, Y., Jones, A., Ndousse, K., Askell, A., Chen, A., DasSarma, N., Drain, D., Fort, S., Ganguli, D., Henighan, T., et~al.
\newblock Training a helpful and harmless assistant with reinforcement learning from human feedback.
\newblock \emph{arXiv preprint arXiv:2204.05862}, 2022{\natexlab{a}}.
\newblock URL \url{https://arxiv.org/abs/2204.05862}.

\bibitem[Bai et~al.(2022{\natexlab{b}})Bai, Kadavath, Kundu, Askell, Kernion, Jones, Chen, Goldie, Mirhoseini, McKinnon, et~al.]{bai2022constitutional}
Bai, Y., Kadavath, S., Kundu, S., Askell, A., Kernion, J., Jones, A., Chen, A., Goldie, A., Mirhoseini, A., McKinnon, C., et~al.
\newblock Constitutional ai: Harmlessness from ai feedback.
\newblock \emph{arXiv preprint arXiv:2212.08073}, 2022{\natexlab{b}}.
\newblock URL \url{https://arxiv.org/abs/2212.08073}.

\bibitem[Ball et~al.(2023)Ball, Smith, Kostrikov, and Levine]{pmlr-v202-ball23a}
Ball, P.~J., Smith, L., Kostrikov, I., and Levine, S.
\newblock Efficient online reinforcement learning with offline data.
\newblock In Krause, A., Brunskill, E., Cho, K., Engelhardt, B., Sabato, S., and Scarlett, J. (eds.), \emph{Proceedings of the 40th International Conference on Machine Learning}, volume 202 of \emph{Proceedings of Machine Learning Research}, pp.\  1577--1594. PMLR, 23--29 Jul 2023.
\newblock URL \url{https://proceedings.mlr.press/v202/ball23a.html}.

\bibitem[Biderman et~al.(2023)Biderman, Schoelkopf, Anthony, Bradley, O'Brien, Hallahan, Khan, Purohit, Prashanth, Raff, Skowron, Sutawika, and Van Der~Wal]{biderman2023pythia}
Biderman, S., Schoelkopf, H., Anthony, Q., Bradley, H., O'Brien, K., Hallahan, E., Khan, M.~A., Purohit, S., Prashanth, U.~S., Raff, E., Skowron, A., Sutawika, L., and Van Der~Wal, O.
\newblock Pythia: a suite for analyzing large language models across training and scaling.
\newblock In \emph{International Conference on Machine Learning \CNFX{ICML}}. JMLR.org, 2023.

\bibitem[Bisk et~al.(2020)Bisk, Zellers, Bras, Gao, and Choi]{bisk2020piqa}
Bisk, Y., Zellers, R., Bras, R.~L., Gao, J., and Choi, Y.
\newblock Piqa: Reasoning about physical commonsense in natural language.
\newblock In \emph{Conference on Artificial Intelligence \CNFX{AAAI}}, 2020.

\bibitem[Bose et~al.(2024)Bose, Xiong, Saha, Du, and Fazel]{bose2024hybridpreferenceoptimizationalignment}
Bose, A., Xiong, Z., Saha, A., Du, S.~S., and Fazel, M.
\newblock Hybrid preference optimization for alignment: Provably faster convergence rates by combining offline preferences with online exploration, 2024.
\newblock URL \url{https://arxiv.org/abs/2412.10616}.

\bibitem[Brown et~al.(2020)Brown, Mann, Ryder, Subbiah, Kaplan, Dhariwal, Neelakantan, Shyam, Sastry, Askell, et~al.]{brown2020language}
Brown, T., Mann, B., Ryder, N., Subbiah, M., Kaplan, J.~D., Dhariwal, P., Neelakantan, A., Shyam, P., Sastry, G., Askell, A., et~al.
\newblock Language models are few-shot learners.
\newblock \emph{Advances in Neural Information Processing Systems \CNFX{NeurIPS}}, 2020.
\newblock URL \url{https://arxiv.org/abs/2005.14165}.

\bibitem[Chiang et~al.(2024{\natexlab{a}})Chiang, Zheng, Sheng, Angelopoulos, Li, Li, Zhang, Zhu, Jordan, Gonzalez, and Stoica]{chiang2024chatbotarenaopenplatform}
Chiang, W.-L., Zheng, L., Sheng, Y., Angelopoulos, A.~N., Li, T., Li, D., Zhang, H., Zhu, B., Jordan, M., Gonzalez, J.~E., and Stoica, I.
\newblock Chatbot arena: An open platform for evaluating llms by human preference, 2024{\natexlab{a}}.
\newblock URL \url{https://arxiv.org/abs/2403.04132}.

\bibitem[Chiang et~al.(2024{\natexlab{b}})Chiang, Zheng, Sheng, Angelopoulos, Li, Li, Zhu, Zhang, Jordan, Gonzalez, et~al.]{chiangchatbot}
Chiang, W.-L., Zheng, L., Sheng, Y., Angelopoulos, A.~N., Li, T., Li, D., Zhu, B., Zhang, H., Jordan, M., Gonzalez, J.~E., et~al.
\newblock Chatbot arena: An open platform for evaluating llms by human preference.
\newblock In \emph{International Conference on Machine Learning \CNFX{ICML}}, 2024{\natexlab{b}}.
\newblock URL \url{https://arxiv.org/pdf/2403.04132}.

\bibitem[Clark et~al.(2018)Clark, Cowhey, Etzioni, Khot, Sabharwal, Schoenick, and Tafjord]{clark2018think}
Clark, P., Cowhey, I., Etzioni, O., Khot, T., Sabharwal, A., Schoenick, C., and Tafjord, O.
\newblock {Think you have Solved Question Answering? Try ARC, the AI2 Reasoning Challenge}.
\newblock \emph{arXiv preprint arXiv:1803.05457}, 2018.
\newblock URL \url{https://arxiv.org/abs/2102.03315}.

\bibitem[Cobbe et~al.(2021)Cobbe, Kosaraju, Bavarian, Chen, Jun, Kaiser, Plappert, Tworek, Hilton, Nakano, Hesse, and Schulman]{cobbe2021gsm8k}
Cobbe, K., Kosaraju, V., Bavarian, M., Chen, M., Jun, H., Kaiser, L., Plappert, M., Tworek, J., Hilton, J., Nakano, R., Hesse, C., and Schulman, J.
\newblock Training verifiers to solve math word problems.
\newblock \emph{arXiv preprint arXiv:2110.14168}, 2021.
\newblock URL \url{https://arxiv.org/pdf/2110.14168}.

\bibitem[Cui et~al.(2024)Cui, Yuan, Ding, Yao, He, Zhu, Ni, Xie, Xie, Lin, Liu, and Sun]{pmlr-v235-cui24f}
Cui, G., Yuan, L., Ding, N., Yao, G., He, B., Zhu, W., Ni, Y., Xie, G., Xie, R., Lin, Y., Liu, Z., and Sun, M.
\newblock {ULTRAFEEDBACK}: Boosting language models with scaled {AI} feedback.
\newblock In Salakhutdinov, R., Kolter, Z., Heller, K., Weller, A., Oliver, N., Scarlett, J., and Berkenkamp, F. (eds.), \emph{Proceedings of the 41st International Conference on Machine Learning}, volume 235 of \emph{Proceedings of Machine Learning Research}, pp.\  9722--9744. PMLR, 21--27 Jul 2024.
\newblock URL \url{https://proceedings.mlr.press/v235/cui24f.html}.

\bibitem[Ding et~al.(2023)Ding, Chen, Xu, Qin, Zheng, Hu, Liu, Sun, and Zhou]{ding2023enhancing}
Ding, N., Chen, Y., Xu, B., Qin, Y., Zheng, Z., Hu, S., Liu, Z., Sun, M., and Zhou, B.
\newblock Enhancing chat language models by scaling high-quality instructional conversations.
\newblock \emph{arXiv preprint 2305.14233}, 2023.
\newblock URL \url{https://arxiv.org/abs/2305.14233}.

\bibitem[Dubey et~al.(2024)Dubey, Jauhri, Pandey, Kadian, Al-Dahle, Letman, Mathur, Schelten, Yang, Fan, Goyal, Hartshorn, Yang, Mitra, Sravankumar, Korenev, Hinsvark, Rao, Zhang, Rodriguez, Gregerson, Spataru, Roziere, Biron, Tang, Chern, Caucheteux, Nayak, Bi, Marra, McConnell, Keller, Touret, Wu, Wong, Ferrer, Nikolaidis, Allonsius, Song, Pintz, Livshits, Esiobu, Choudhary, Mahajan, Garcia-Olano, Perino, Hupkes, Lakomkin, AlBadawy, Lobanova, Dinan, Smith, Radenovic, Zhang, Synnaeve, Lee, Anderson, Nail, Mialon, Pang, Cucurell, Nguyen, Korevaar, Xu, Touvron, Zarov, Ibarra, Kloumann, Misra, Evtimov, Copet, Lee, Geffert, Vranes, Park, Mahadeokar, Shah, van~der Linde, Billock, Hong, Lee, Fu, Chi, Huang, Liu, Wang, Yu, Bitton, Spisak, Park, Rocca, Johnstun, Saxe, Jia, Alwala, Upasani, Plawiak, Li, Heafield, Stone, El-Arini, Iyer, Malik, Chiu, Bhalla, Rantala-Yeary, van~der Maaten, Chen, Tan, Jenkins, Martin, Madaan, Malo, Blecher, Landzaat, de~Oliveira, Muzzi, Pasupuleti, Singh, Paluri, Kardas, Oldham, Rita,
  Pavlova, Kambadur, Lewis, Si, Singh, Hassan, Goyal, Torabi, Bashlykov, Bogoychev, Chatterji, Duchenne, \c{C}elebi, Alrassy, Zhang, Li, Vasic, Weng, Bhargava, Dubal, Krishnan, Koura, Xu, He, Dong, Srinivasan, Ganapathy, Calderer, Cabral, Stojnic, Raileanu, Girdhar, Patel, Sauvestre, Polidoro, Sumbaly, Taylor, Silva, Hou, Wang, Hosseini, Chennabasappa, Singh, Bell, Kim, Edunov, Nie, Narang, Raparthy, Shen, Wan, Bhosale, Zhang, Vandenhende, Batra, Whitman, Sootla, Collot, Gururangan, Borodinsky, Herman, Fowler, Sheasha, Georgiou, Scialom, Speckbacher, Mihaylov, Xiao, Karn, Goswami, Gupta, Ramanathan, Kerkez, Gonguet, Do, Vogeti, Petrovic, Chu, Xiong, Fu, Meers, Martinet, Wang, Tan, Xie, Jia, Wang, Goldschlag, Gaur, Babaei, Wen, Song, Zhang, Li, Mao, Coudert, Yan, Chen, Papakipos, Singh, Grattafiori, Jain, Kelsey, Shajnfeld, Gangidi, Victoria, Goldstand, Menon, Sharma, Boesenberg, Vaughan, Baevski, Feinstein, Kallet, Sangani, Yunus, Lupu, Alvarado, Caples, Gu, Ho, Poulton, Ryan, Ramchandani, Franco, Saraf,
  Chowdhury, Gabriel, Bharambe, Eisenman, Yazdan, James, Maurer, Leonhardi, Huang, Loyd, Paola, Paranjape, Liu, Wu, Ni, Hancock, Wasti, Spence, Stojkovic, Gamido, Montalvo, Parker, Burton, Mejia, Wang, Kim, Zhou, Hu, Chu, Cai, Tindal, Feichtenhofer, Civin, Beaty, Kreymer, Li, Wyatt, Adkins, Xu, Testuggine, David, Parikh, Liskovich, Foss, Wang, Le, Holland, Dowling, Jamil, Montgomery, Presani, Hahn, Wood, Brinkman, Arcaute, Dunbar, Smothers, Sun, Kreuk, Tian, Ozgenel, Caggioni, Guzm\'{a}n, Kanayet, Seide, Florez, Schwarz, Badeer, Swee, Halpern, Thattai, Herman, Sizov, Guangyi, Zhang, Lakshminarayanan, Shojanazeri, Zou, Wang, Zha, Habeeb, Rudolph, Suk, Aspegren, Goldman, Damlaj, Molybog, Tufanov, Veliche, Gat, Weissman, Geboski, Kohli, Asher, Gaya, Marcus, Tang, Chan, Zhen, Reizenstein, Teboul, Zhong, Jin, Yang, Cummings, Carvill, Shepard, McPhie, Torres, Ginsburg, Wang, Wu, U, Saxena, Prasad, Khandelwal, Zand, Matosich, Veeraraghavan, Michelena, Li, Huang, Chawla, Lakhotia, Huang, Chen, Garg, A, Silva, Bell,
  Zhang, Guo, Yu, Moshkovich, Wehrstedt, Khabsa, Avalani, Bhatt, Tsimpoukelli, Mankus, Hasson, Lennie, Reso, Groshev, Naumov, Lathi, Keneally, Seltzer, Valko, Restrepo, Patel, Vyatskov, Samvelyan, Clark, Macey, Wang, Hermoso, Metanat, Rastegari, Bansal, Santhanam, Parks, White, Bawa, Singhal, Egebo, Usunier, Laptev, Dong, Zhang, Cheng, Chernoguz, Hart, Salpekar, Kalinli, Kent, Parekh, Saab, Balaji, Rittner, Bontrager, Roux, Dollar, Zvyagina, Ratanchandani, Yuvraj, Liang, Alao, Rodriguez, Ayub, Murthy, Nayani, Mitra, Li, Hogan, Battey, Wang, Maheswari, Howes, Rinott, Bondu, Datta, Chugh, Hunt, Dhillon, Sidorov, Pan, Verma, Yamamoto, Ramaswamy, Lindsay, Lindsay, Feng, Lin, Zha, Shankar, Zhang, Zhang, Wang, Agarwal, Sajuyigbe, Chintala, Max, Chen, Kehoe, Satterfield, Govindaprasad, Gupta, Cho, Virk, Subramanian, Choudhury, Goldman, Remez, Glaser, Best, Kohler, Robinson, Li, Zhang, Matthews, Chou, Shaked, Vontimitta, Ajayi, Montanez, Mohan, Kumar, Mangla, Albiero, Ionescu, Poenaru, Mihailescu, Ivanov, Li, Wang,
  Jiang, Bouaziz, Constable, Tang, Wang, Wu, Wang, Xia, Wu, Gao, Chen, Hu, Jia, Qi, Li, Zhang, Zhang, Adi, Nam, Yu, Wang, Hao, Qian, He, Rait, DeVito, Rosnbrick, Wen, Yang, and Zhao]{dubey2024llama3herdmodels}
Dubey, A., Jauhri, A., Pandey, A., Kadian, A., Al-Dahle, A., Letman, A., Mathur, A., Schelten, A., Yang, A., Fan, A., Goyal, A., Hartshorn, A., Yang, A., Mitra, A., Sravankumar, A., Korenev, A., Hinsvark, A., Rao, A., Zhang, A., Rodriguez, A., Gregerson, A., Spataru, A., Roziere, B., Biron, B., Tang, B., Chern, B., Caucheteux, C., Nayak, C., Bi, C., Marra, C., McConnell, C., Keller, C., Touret, C., Wu, C., Wong, C., Ferrer, C.~C., Nikolaidis, C., Allonsius, D., Song, D., Pintz, D., Livshits, D., Esiobu, D., Choudhary, D., Mahajan, D., Garcia-Olano, D., Perino, D., Hupkes, D., Lakomkin, E., AlBadawy, E., Lobanova, E., Dinan, E., Smith, E.~M., Radenovic, F., Zhang, F., Synnaeve, G., Lee, G., Anderson, G.~L., Nail, G., Mialon, G., Pang, G., Cucurell, G., Nguyen, H., Korevaar, H., Xu, H., Touvron, H., Zarov, I., Ibarra, I.~A., Kloumann, I., Misra, I., Evtimov, I., Copet, J., Lee, J., Geffert, J., Vranes, J., Park, J., Mahadeokar, J., Shah, J., van~der Linde, J., Billock, J., Hong, J., Lee, J., Fu, J., Chi, J.,
  Huang, J., Liu, J., Wang, J., Yu, J., Bitton, J., Spisak, J., Park, J., Rocca, J., Johnstun, J., Saxe, J., Jia, J., Alwala, K.~V., Upasani, K., Plawiak, K., Li, K., Heafield, K., Stone, K., El-Arini, K., Iyer, K., Malik, K., Chiu, K., Bhalla, K., Rantala-Yeary, L., van~der Maaten, L., Chen, L., Tan, L., Jenkins, L., Martin, L., Madaan, L., Malo, L., Blecher, L., Landzaat, L., de~Oliveira, L., Muzzi, M., Pasupuleti, M., Singh, M., Paluri, M., Kardas, M., Oldham, M., Rita, M., Pavlova, M., Kambadur, M., Lewis, M., Si, M., Singh, M.~K., Hassan, M., Goyal, N., Torabi, N., Bashlykov, N., Bogoychev, N., Chatterji, N., Duchenne, O., \c{C}elebi, O., Alrassy, P., Zhang, P., Li, P., Vasic, P., Weng, P., Bhargava, P., Dubal, P., Krishnan, P., Koura, P.~S., Xu, P., He, Q., Dong, Q., Srinivasan, R., Ganapathy, R., Calderer, R., Cabral, R.~S., Stojnic, R., Raileanu, R., Girdhar, R., Patel, R., Sauvestre, R., Polidoro, R., Sumbaly, R., Taylor, R., Silva, R., Hou, R., Wang, R., Hosseini, S., Chennabasappa, S., Singh, S.,
  Bell, S., Kim, S.~S., Edunov, S., Nie, S., Narang, S., Raparthy, S., Shen, S., Wan, S., Bhosale, S., Zhang, S., Vandenhende, S., Batra, S., Whitman, S., Sootla, S., Collot, S., Gururangan, S., Borodinsky, S., Herman, T., Fowler, T., Sheasha, T., Georgiou, T., Scialom, T., Speckbacher, T., Mihaylov, T., Xiao, T., Karn, U., Goswami, V., Gupta, V., Ramanathan, V., Kerkez, V., Gonguet, V., Do, V., Vogeti, V., Petrovic, V., Chu, W., Xiong, W., Fu, W., Meers, W., Martinet, X., Wang, X., Tan, X.~E., Xie, X., Jia, X., Wang, X., Goldschlag, Y., Gaur, Y., Babaei, Y., Wen, Y., Song, Y., Zhang, Y., Li, Y., Mao, Y., Coudert, Z.~D., Yan, Z., Chen, Z., Papakipos, Z., Singh, A., Grattafiori, A., Jain, A., Kelsey, A., Shajnfeld, A., Gangidi, A., Victoria, A., Goldstand, A., Menon, A., Sharma, A., Boesenberg, A., Vaughan, A., Baevski, A., Feinstein, A., Kallet, A., Sangani, A., Yunus, A., Lupu, A., Alvarado, A., Caples, A., Gu, A., Ho, A., Poulton, A., Ryan, A., Ramchandani, A., Franco, A., Saraf, A., Chowdhury, A., Gabriel,
  A., Bharambe, A., Eisenman, A., Yazdan, A., James, B., Maurer, B., Leonhardi, B., Huang, B., Loyd, B., Paola, B.~D., Paranjape, B., Liu, B., Wu, B., Ni, B., Hancock, B., Wasti, B., Spence, B., Stojkovic, B., Gamido, B., Montalvo, B., Parker, C., Burton, C., Mejia, C., Wang, C., Kim, C., Zhou, C., Hu, C., Chu, C.-H., Cai, C., Tindal, C., Feichtenhofer, C., Civin, D., Beaty, D., Kreymer, D., Li, D., Wyatt, D., Adkins, D., Xu, D., Testuggine, D., David, D., Parikh, D., Liskovich, D., Foss, D., Wang, D., Le, D., Holland, D., Dowling, E., Jamil, E., Montgomery, E., Presani, E., Hahn, E., Wood, E., Brinkman, E., Arcaute, E., Dunbar, E., Smothers, E., Sun, F., Kreuk, F., Tian, F., Ozgenel, F., Caggioni, F., Guzm\'{a}n, F., Kanayet, F., Seide, F., Florez, G.~M., Schwarz, G., Badeer, G., Swee, G., Halpern, G., Thattai, G., Herman, G., Sizov, G., Guangyi, Zhang, Lakshminarayanan, G., Shojanazeri, H., Zou, H., Wang, H., Zha, H., Habeeb, H., Rudolph, H., Suk, H., Aspegren, H., Goldman, H., Damlaj, I., Molybog, I.,
  Tufanov, I., Veliche, I.-E., Gat, I., Weissman, J., Geboski, J., Kohli, J., Asher, J., Gaya, J.-B., Marcus, J., Tang, J., Chan, J., Zhen, J., Reizenstein, J., Teboul, J., Zhong, J., Jin, J., Yang, J., Cummings, J., Carvill, J., Shepard, J., McPhie, J., Torres, J., Ginsburg, J., Wang, J., Wu, K., U, K.~H., Saxena, K., Prasad, K., Khandelwal, K., Zand, K., Matosich, K., Veeraraghavan, K., Michelena, K., Li, K., Huang, K., Chawla, K., Lakhotia, K., Huang, K., Chen, L., Garg, L., A, L., Silva, L., Bell, L., Zhang, L., Guo, L., Yu, L., Moshkovich, L., Wehrstedt, L., Khabsa, M., Avalani, M., Bhatt, M., Tsimpoukelli, M., Mankus, M., Hasson, M., Lennie, M., Reso, M., Groshev, M., Naumov, M., Lathi, M., Keneally, M., Seltzer, M.~L., Valko, M., Restrepo, M., Patel, M., Vyatskov, M., Samvelyan, M., Clark, M., Macey, M., Wang, M., Hermoso, M.~J., Metanat, M., Rastegari, M., Bansal, M., Santhanam, N., Parks, N., White, N., Bawa, N., Singhal, N., Egebo, N., Usunier, N., Laptev, N.~P., Dong, N., Zhang, N., Cheng, N.,
  Chernoguz, O., Hart, O., Salpekar, O., Kalinli, O., Kent, P., Parekh, P., Saab, P., Balaji, P., Rittner, P., Bontrager, P., Roux, P., Dollar, P., Zvyagina, P., Ratanchandani, P., Yuvraj, P., Liang, Q., Alao, R., Rodriguez, R., Ayub, R., Murthy, R., Nayani, R., Mitra, R., Li, R., Hogan, R., Battey, R., Wang, R., Maheswari, R., Howes, R., Rinott, R., Bondu, S.~J., Datta, S., Chugh, S., Hunt, S., Dhillon, S., Sidorov, S., Pan, S., Verma, S., Yamamoto, S., Ramaswamy, S., Lindsay, S., Lindsay, S., Feng, S., Lin, S., Zha, S.~C., Shankar, S., Zhang, S., Zhang, S., Wang, S., Agarwal, S., Sajuyigbe, S., Chintala, S., Max, S., Chen, S., Kehoe, S., Satterfield, S., Govindaprasad, S., Gupta, S., Cho, S., Virk, S., Subramanian, S., Choudhury, S., Goldman, S., Remez, T., Glaser, T., Best, T., Kohler, T., Robinson, T., Li, T., Zhang, T., Matthews, T., Chou, T., Shaked, T., Vontimitta, V., Ajayi, V., Montanez, V., Mohan, V., Kumar, V.~S., Mangla, V., Albiero, V., Ionescu, V., Poenaru, V., Mihailescu, V.~T., Ivanov, V., Li,
  W., Wang, W., Jiang, W., Bouaziz, W., Constable, W., Tang, X., Wang, X., Wu, X., Wang, X., Xia, X., Wu, X., Gao, X., Chen, Y., Hu, Y., Jia, Y., Qi, Y., Li, Y., Zhang, Y., Zhang, Y., Adi, Y., Nam, Y., Yu, Wang, Hao, Y., Qian, Y., He, Y., Rait, Z., DeVito, Z., Rosnbrick, Z., Wen, Z., Yang, Z., and Zhao, Z.
\newblock The llama 3 herd of models, 2024.
\newblock URL \url{https://arxiv.org/abs/2407.21783}.

\bibitem[Dubois et~al.(2023)Dubois, Li, Taori, Zhang, Gulrajani, Ba, Guestrin, Liang, and Hashimoto]{dubois2023alpacafarm}
Dubois, Y., Li, X., Taori, R., Zhang, T., Gulrajani, I., Ba, J., Guestrin, C., Liang, P., and Hashimoto, T.
\newblock Alpacafarm: A simulation framework for methods that learn from human feedback.
\newblock In \emph{Thirty-seventh Conference on Neural Information Processing Systems}, 2023.
\newblock URL \url{https://openreview.net/forum?id=4hturzLcKX}.

\bibitem[Dubois et~al.(2024)Dubois, Liang, and Hashimoto]{dubois2024lengthcontrolled}
Dubois, Y., Liang, P., and Hashimoto, T.
\newblock Length-controlled alpacaeval: A simple debiasing of automatic evaluators.
\newblock In \emph{First Conference on Language Modeling}, 2024.
\newblock URL \url{https://openreview.net/forum?id=CybBmzWBX0}.

\bibitem[Ethayarajh et~al.(2024)Ethayarajh, Xu, Muennighoff, Jurafsky, and Kiela]{ethayarajhmodel}
Ethayarajh, K., Xu, W., Muennighoff, N., Jurafsky, D., and Kiela, D.
\newblock Model alignment as prospect theoretic optimization.
\newblock In \emph{International Conference on Machine Learning \CNFX{ICML}}, 2024.

\bibitem[Gao et~al.(2024)Gao, Tow, Abbasi, Biderman, Black, DiPofi, Foster, Golding, Hsu, Le~Noac'h, Li, McDonell, Muennighoff, Ociepa, Phang, Reynolds, Schoelkopf, Skowron, Sutawika, Tang, Thite, Wang, Wang, and Zou]{eval-harness}
Gao, L., Tow, J., Abbasi, B., Biderman, S., Black, S., DiPofi, A., Foster, C., Golding, L., Hsu, J., Le~Noac'h, A., Li, H., McDonell, K., Muennighoff, N., Ociepa, C., Phang, J., Reynolds, L., Schoelkopf, H., Skowron, A., Sutawika, L., Tang, E., Thite, A., Wang, B., Wang, K., and Zou, A.
\newblock A framework for few-shot language model evaluation, 07 2024.
\newblock URL \url{https://zenodo.org/records/12608602}.

\bibitem[Gu et~al.(2017)Gu, Lillicrap, Ghahramani, Turner, Sch\"{o}lkopf, and Levine]{10.5555/3294996.3295141}
Gu, S., Lillicrap, T., Ghahramani, Z., Turner, R.~E., Sch\"{o}lkopf, B., and Levine, S.
\newblock Interpolated policy gradient: merging on-policy and off-policy gradient estimation for deep reinforcement learning.
\newblock In \emph{Advances in Neural Information Processing Systems \CNFX{NeurIPS}}, NIPS'17, pp.\  3849–3858, Red Hook, NY, USA, 2017. Curran Associates Inc.
\newblock ISBN 9781510860964.

\bibitem[Guo et~al.(2024)Guo, Zhang, Liu, Liu, Khalman, Llinares, Rame, Mesnard, Zhao, Piot, Ferret, and Blondel]{guo2024directlanguagemodelalignment}
Guo, S., Zhang, B., Liu, T., Liu, T., Khalman, M., Llinares, F., Rame, A., Mesnard, T., Zhao, Y., Piot, B., Ferret, J., and Blondel, M.
\newblock Direct language model alignment from online ai feedback, 2024.
\newblock URL \url{https://arxiv.org/abs/2402.04792}.

\bibitem[Hendrycks et~al.(2020)Hendrycks, Burns, Basart, Zou, Mazeika, Song, and Steinhardt]{hendrycks2020measuring}
Hendrycks, D., Burns, C., Basart, S., Zou, A., Mazeika, M., Song, D., and Steinhardt, J.
\newblock Measuring massive multitask language understanding.
\newblock In \emph{International Conference on Learning Representations \CNFX{ICLR}}, 2020.

\bibitem[Hu et~al.(2024)Hu, Wu, Zhu, Xianyu, Wang, Zhang, and Cao]{hu2024openrlhf}
Hu, J., Wu, X., Zhu, Z., Xianyu, Wang, W., Zhang, D., and Cao, Y.
\newblock Openrlhf: An easy-to-use, scalable and high-performance rlhf framework.
\newblock \emph{arXiv preprint arXiv:2405.11143}, 2024.

\bibitem[Hu et~al.(2023)Hu, Luo, Wang, Cheng, Liu, and Sun]{hu-etal-2023-wont}
Hu, S., Luo, Y., Wang, H., Cheng, X., Liu, Z., and Sun, M.
\newblock Won`t get fooled again: Answering questions with false premises.
\newblock In Rogers, A., Boyd-Graber, J., and Okazaki, N. (eds.), \emph{Proceedings of the 61st Annual Meeting of the Association for Computational Linguistics (Volume 1: Long Papers)}, pp.\  5626--5643, Toronto, Canada, July 2023. Association for Computational Linguistics.
\newblock \doi{10.18653/v1/2023.acl-long.309}.
\newblock URL \url{https://aclanthology.org/2023.acl-long.309/}.

\bibitem[Ivison et~al.(2024)Ivison, Wang, Liu, Wu, Pyatkin, Lambert, Smith, Choi, and Hajishirzi]{ivison2024unpacking}
Ivison, H., Wang, Y., Liu, J., Wu, Z., Pyatkin, V., Lambert, N., Smith, N.~A., Choi, Y., and Hajishirzi, H.
\newblock Unpacking dpo and ppo: Disentangling best practices for learning from preference feedback.
\newblock In \emph{Advances in Neural Information Processing Systems \CNFX{NeurIPS}}, 2024.

\bibitem[Jiang et~al.(2023)Jiang, Sablayrolles, Mensch, Bamford, Chaplot, de~Las~Casas, Bressand, Lengyel, Lample, Saulnier, Lavaud, Lachaux, Stock, Scao, Lavril, Wang, Lacroix, and Sayed]{Jiang2023Mistral7}
Jiang, A.~Q., Sablayrolles, A., Mensch, A., Bamford, C., Chaplot, D.~S., de~Las~Casas, D., Bressand, F., Lengyel, G., Lample, G., Saulnier, L., Lavaud, L.~R., Lachaux, M.-A., Stock, P., Scao, T.~L., Lavril, T., Wang, T., Lacroix, T., and Sayed, W.~E.
\newblock Mistral 7b.
\newblock \emph{ArXiv}, abs/2310.06825, 2023.
\newblock URL \url{https://api.semanticscholar.org/CorpusID:263830494}.

\bibitem[Kim et~al.(2024)Kim, Goyal, Zhang, Xiong, Hou, Kambadur, Mahajan, Hajishirzi, and Tan]{kim2024systematicexaminationpreferencelearning}
Kim, J., Goyal, A., Zhang, A., Xiong, B., Hou, R., Kambadur, M., Mahajan, D., Hajishirzi, H., and Tan, L.
\newblock A systematic examination of preference learning through the lens of instruction-following, 2024.
\newblock URL \url{https://arxiv.org/abs/2412.15282}.

\bibitem[Knight et~al.(2017)Knight, {Badarau, Bianca}, {Baranescu, Laura}, {Bonial, Claire}, {Bardocz, Madalina}, {Griffitt, Kira}, {Hermjakob, Ulf}, {Marcu, Daniel}, {Palmer, Martha}, {O'Gorman, Tim}, and {Schneider, Nathan}]{amr2.0-https://doi.org/10.35111/s444-np87}
Knight, K., {Badarau, Bianca}, {Baranescu, Laura}, {Bonial, Claire}, {Bardocz, Madalina}, {Griffitt, Kira}, {Hermjakob, Ulf}, {Marcu, Daniel}, {Palmer, Martha}, {O'Gorman, Tim}, and {Schneider, Nathan}.
\newblock Abstract meaning representation (amr) annotation release 2.0, 2017.
\newblock URL \url{https://catalog.ldc.upenn.edu/LDC2017T10}.

\bibitem[Lambert et~al.(2024{\natexlab{a}})Lambert, Morrison, Pyatkin, Huang, Ivison, Brahman, Miranda, Liu, Dziri, Lyu, Gu, Malik, Graf, Hwang, Yang, Bras, Tafjord, Wilhelm, Soldaini, Smith, Wang, Dasigi, and Hajishirzi]{lambert2024tulu3pushingfrontiers}
Lambert, N., Morrison, J., Pyatkin, V., Huang, S., Ivison, H., Brahman, F., Miranda, L. J.~V., Liu, A., Dziri, N., Lyu, S., Gu, Y., Malik, S., Graf, V., Hwang, J.~D., Yang, J., Bras, R.~L., Tafjord, O., Wilhelm, C., Soldaini, L., Smith, N.~A., Wang, Y., Dasigi, P., and Hajishirzi, H.
\newblock T\"ulu 3: Pushing frontiers in open language model post-training, 2024{\natexlab{a}}.
\newblock URL \url{https://arxiv.org/abs/2411.15124}.

\bibitem[Lambert et~al.(2024{\natexlab{b}})Lambert, Pyatkin, Morrison, Miranda, Lin, Chandu, Dziri, Kumar, Zick, Choi, Smith, and Hajishirzi]{lambert2024rewardbench}
Lambert, N., Pyatkin, V., Morrison, J., Miranda, L., Lin, B.~Y., Chandu, K., Dziri, N., Kumar, S., Zick, T., Choi, Y., Smith, N.~A., and Hajishirzi, H.
\newblock Rewardbench: Evaluating reward models for language modeling, 2024{\natexlab{b}}.

\bibitem[Lee et~al.(2021)Lee, Seo, Lee, Abbeel, and Shin]{lee2021offlinetoonline}
Lee, S., Seo, Y., Lee, K., Abbeel, P., and Shin, J.
\newblock Offline-to-online reinforcement learning via balanced replay and pessimistic q-ensemble.
\newblock In \emph{5th Annual Conference on Robot Learning}, 2021.
\newblock URL \url{https://openreview.net/forum?id=AlJXhEI6J5W}.

\bibitem[Li et~al.(2023)Li, Zhang, Dubois, Taori, Gulrajani, Guestrin, Liang, and Hashimoto]{alpaca_eval}
Li, X., Zhang, T., Dubois, Y., Taori, R., Gulrajani, I., Guestrin, C., Liang, P., and Hashimoto, T.~B.
\newblock Alpacaeval: An automatic evaluator of instruction-following models.
\newblock \url{https://github.com/tatsu-lab/alpaca_eval}, 5 2023.

\bibitem[Lin et~al.(2021)Lin, Hilton, and Evans]{lin2021truthfulqa}
Lin, S., Hilton, J., and Evans, O.
\newblock {TruthfulQA:} measuring how models mimic human falsehoods.
\newblock \emph{arXiv preprint arXiv:2109.07958}, 2021.
\newblock URL \url{https://arxiv.org/abs/2109.07958}.

\bibitem[Longpre et~al.(2023)Longpre, Hou, Vu, Webson, Chung, Tay, Zhou, Le, Zoph, Wei, et~al.]{longpre2023flan}
Longpre, S., Hou, L., Vu, T., Webson, A., Chung, H.~W., Tay, Y., Zhou, D., Le, Q.~V., Zoph, B., Wei, J., et~al.
\newblock The flan collection: Designing data and methods for effective instruction tuning.
\newblock \emph{arXiv preprint arXiv:2301.13688}, 2023.
\newblock URL \url{https://arxiv.org/abs/2301.13688}.

\bibitem[Lu et~al.(2024)Lu, Wang, Li, Jiang, and Khashabi]{lu2024benchmarkinglanguagemodelcreativity}
Lu, Y., Wang, D., Li, T., Jiang, D., and Khashabi, D.
\newblock Benchmarking language model creativity: A case study on code generation.
\newblock \emph{arXiv preprint arXiv:2407.09007}, 2024.
\newblock URL \url{https://arxiv.org/abs/2407.09007}.

\bibitem[Meng et~al.(2024)Meng, Xia, and Chen]{meng2024simpo}
Meng, Y., Xia, M., and Chen, D.
\newblock Simpo: Simple preference optimization with a reference-free reward.
\newblock In \emph{Advances in Neural Information Processing Systems \CNFX{NeurIPS}}, 2024.

\bibitem[Mihaylov et~al.(2018{\natexlab{a}})Mihaylov, Clark, Khot, and Sabharwal]{OpenBookQA2018}
Mihaylov, T., Clark, P., Khot, T., and Sabharwal, A.
\newblock Can a suit of armor conduct electricity? a new dataset for open book question answering.
\newblock In \emph{EMNLP}, 2018{\natexlab{a}}.

\bibitem[Mihaylov et~al.(2018{\natexlab{b}})Mihaylov, Clark, Khot, and Sabharwal]{mihaylov2018can}
Mihaylov, T., Clark, P., Khot, T., and Sabharwal, A.
\newblock Can a suit of armor conduct electricity? a new dataset for open book question answering.
\newblock In \emph{Conference on Empirical Methods in Natural Language Processing \CNFX{EMNLP}}, 2018{\natexlab{b}}.
\newblock URL \url{https://arxiv.org/abs/1809.02789}.

\bibitem[Nakamoto et~al.(2023)Nakamoto, Zhai, Singh, Mark, Ma, Finn, Kumar, and Levine]{nakamoto2023calql}
Nakamoto, M., Zhai, Y., Singh, A., Mark, M.~S., Ma, Y., Finn, C., Kumar, A., and Levine, S.
\newblock Cal-{QL}: Calibrated offline {RL} pre-training for efficient online fine-tuning.
\newblock In \emph{Advances in Neural Information Processing Systems \CNFX{NeurIPS}}, 2023.
\newblock URL \url{https://openreview.net/forum?id=GcEIvidYSw}.

\bibitem[Nvidia et~al.(2024)Nvidia, :, Adler, Agarwal, Aithal, Anh, Bhattacharya, Brundyn, Casper, Catanzaro, Clay, Cohen, Das, Dattagupta, Delalleau, Derczynski, Dong, Egert, Evans, Ficek, Fridman, Ghosh, Ginsburg, Gitman, Grzegorzek, Hero, Huang, Jawa, Jennings, Jhunjhunwala, Kamalu, Khan, Kuchaiev, LeGresley, Li, Liu, Liu, Long, Mahabaleshwarkar, Majumdar, Maki, Martinez, de~Melo, Moshkov, Narayanan, Narenthiran, Navarro, Nguyen, Nitski, Noroozi, Nutheti, Parisien, Parmar, Patwary, Pawelec, Ping, Prabhumoye, Roy, Saar, Sabavat, Satheesh, Scowcroft, Sewall, Shamis, Shen, Shoeybi, Sizer, Smelyanskiy, Soares, Sreedhar, Su, Subramanian, Sun, Toshniwal, Wang, Wang, You, Zeng, Zhang, Zhang, Zhang, Zhang, and Zhu]{nvidia2024nemotron4340btechnicalreport}
Nvidia, :, Adler, B., Agarwal, N., Aithal, A., Anh, D.~H., Bhattacharya, P., Brundyn, A., Casper, J., Catanzaro, B., Clay, S., Cohen, J., Das, S., Dattagupta, A., Delalleau, O., Derczynski, L., Dong, Y., Egert, D., Evans, E., Ficek, A., Fridman, D., Ghosh, S., Ginsburg, B., Gitman, I., Grzegorzek, T., Hero, R., Huang, J., Jawa, V., Jennings, J., Jhunjhunwala, A., Kamalu, J., Khan, S., Kuchaiev, O., LeGresley, P., Li, H., Liu, J., Liu, Z., Long, E., Mahabaleshwarkar, A.~S., Majumdar, S., Maki, J., Martinez, M., de~Melo, M.~R., Moshkov, I., Narayanan, D., Narenthiran, S., Navarro, J., Nguyen, P., Nitski, O., Noroozi, V., Nutheti, G., Parisien, C., Parmar, J., Patwary, M., Pawelec, K., Ping, W., Prabhumoye, S., Roy, R., Saar, T., Sabavat, V. R.~N., Satheesh, S., Scowcroft, J.~P., Sewall, J., Shamis, P., Shen, G., Shoeybi, M., Sizer, D., Smelyanskiy, M., Soares, F., Sreedhar, M.~N., Su, D., Subramanian, S., Sun, S., Toshniwal, S., Wang, H., Wang, Z., You, J., Zeng, J., Zhang, J., Zhang, J., Zhang, V., Zhang, Y.,
  and Zhu, C.
\newblock Nemotron-4 340b technical report, 2024.
\newblock URL \url{https://arxiv.org/abs/2406.11704}.

\bibitem[Ouyang et~al.(2022)Ouyang, Wu, Jiang, Almeida, Wainwright, Mishkin, Zhang, Agarwal, Slama, Ray, et~al.]{ouyang2022training}
Ouyang, L., Wu, J., Jiang, X., Almeida, D., Wainwright, C.~L., Mishkin, P., Zhang, C., Agarwal, S., Slama, K., Ray, A., et~al.
\newblock {Training Language Models to Follow Instructions with Human Feedback}.
\newblock In \emph{Advances in Neural Information Processing Systems \CNFX{NeurIPS}}, 2022.
\newblock URL \url{https://arxiv.org/abs/2203.02155}.

\bibitem[Pal et~al.(2024)Pal, Karkhanis, Dooley, Roberts, Naidu, and White]{pal2024smaugfixingfailuremodes}
Pal, A., Karkhanis, D., Dooley, S., Roberts, M., Naidu, S., and White, C.
\newblock Smaug: Fixing failure modes of preference optimisation with dpo-positive, 2024.
\newblock URL \url{https://arxiv.org/abs/2402.13228}.

\bibitem[Panickssery et~al.(2024)Panickssery, Bowman, and Feng]{panickssery2024llm}
Panickssery, A., Bowman, S.~R., and Feng, S.
\newblock {LLM} evaluators recognize and favor their own generations.
\newblock In \emph{The Thirty-eighth Annual Conference on Neural Information Processing Systems}, 2024.
\newblock URL \url{https://openreview.net/forum?id=4NJBV6Wp0h}.

\bibitem[Park et~al.(2024)Park, Jwa, Meiying, Kim, and Choi]{park-etal-2024-offsetbias}
Park, J., Jwa, S., Meiying, R., Kim, D., and Choi, S.
\newblock {O}ffset{B}ias: Leveraging debiased data for tuning evaluators.
\newblock In Al-Onaizan, Y., Bansal, M., and Chen, Y.-N. (eds.), \emph{Findings of the Association for Computational Linguistics: EMNLP 2024}, pp.\  1043--1067, Miami, Florida, USA, November 2024. Association for Computational Linguistics.
\newblock \doi{10.18653/v1/2024.findings-emnlp.57}.
\newblock URL \url{https://aclanthology.org/2024.findings-emnlp.57/}.

\bibitem[Penedo et~al.(2024)Penedo, Kydl\'{\i}\v{c}ek, allal, Lozhkov, Mitchell, Raffel, Werra, and Wolf]{penedo2024finewebdatasetsdecantingweb}
Penedo, G., Kydl\'{\i}\v{c}ek, H., allal, L.~B., Lozhkov, A., Mitchell, M., Raffel, C., Werra, L.~V., and Wolf, T.
\newblock The fineweb datasets: Decanting the web for the finest text data at scale, 2024.
\newblock URL \url{https://arxiv.org/abs/2406.17557}.

\bibitem[Rafailov et~al.(2024)Rafailov, Sharma, Mitchell, Manning, Ermon, and Finn]{rafailov2024direct}
Rafailov, R., Sharma, A., Mitchell, E., Manning, C.~D., Ermon, S., and Finn, C.
\newblock Direct preference optimization: Your language model is secretly a reward model.
\newblock \emph{Advances in Neural Information Processing Systems \CNFX{NeurIPS}}, 36, 2024.
\newblock URL \url{https://arxiv.org/abs/2305.18290}.

\bibitem[Rashidinejad \& Tian(2024)Rashidinejad and Tian]{rashidinejad2024sailheadwindalignmentrobust}
Rashidinejad, P. and Tian, Y.
\newblock Sail into the headwind: Alignment via robust rewards and dynamic labels against reward hacking, 2024.
\newblock URL \url{https://arxiv.org/abs/2412.09544}.

\bibitem[Razin et~al.(2024)Razin, Malladi, Bhaskar, Chen, Arora, and Hanin]{razin2024unintentional}
Razin, N., Malladi, S., Bhaskar, A., Chen, D., Arora, S., and Hanin, B.
\newblock Unintentional unalignment: Likelihood displacement in direct preference optimization.
\newblock \emph{arXiv preprint arXiv:2410.08847}, 2024.

\bibitem[Sakaguchi et~al.(2020)Sakaguchi, Bras, Bhagavatula, and Choi]{sakaguchi2020winogrande}
Sakaguchi, K., Bras, R.~L., Bhagavatula, C., and Choi, Y.
\newblock {WINOGRANDE:} an adversarial winograd schema challenge at scale.
\newblock In \emph{Conference on Artificial Intelligence \CNFX{AAAI}}, 2020.
\newblock URL \url{https://arxiv.org/abs/1907.10641}.

\bibitem[Schulman et~al.(2017)Schulman, Wolski, Dhariwal, Radford, and Klimov]{schulman2017proximal}
Schulman, J., Wolski, F., Dhariwal, P., Radford, A., and Klimov, O.
\newblock Proximal policy optimization algorithms.
\newblock \emph{arXiv preprint arXiv:1707.06347}, 2017.
\newblock URL \url{https://arxiv.org/abs/1707.06347}.

\bibitem[Shi et~al.(2024)Shi, Zhou, and Du]{shi2024crucialrolesamplersonline}
Shi, R., Zhou, R., and Du, S.~S.
\newblock The crucial role of samplers in online direct preference optimization, 2024.
\newblock URL \url{https://arxiv.org/abs/2409.19605}.

\bibitem[Singhal et~al.(2024)Singhal, Goyal, Xu, and Durrett]{singhal2024a}
Singhal, P., Goyal, T., Xu, J., and Durrett, G.
\newblock A long way to go: Investigating length correlations in {RLHF}.
\newblock In \emph{First Conference on Language Modeling}, 2024.
\newblock URL \url{https://openreview.net/forum?id=G8LaO1P0xv}.

\bibitem[Song et~al.(2023)Song, Zhou, Sekhari, Bagnell, Krishnamurthy, and Sun]{song2023hybrid}
Song, Y., Zhou, Y., Sekhari, A., Bagnell, D., Krishnamurthy, A., and Sun, W.
\newblock Hybrid {RL}: Using both offline and online data can make {RL} efficient.
\newblock In \emph{International Conference on Learning Representations \CNFX{ICLR}}, 2023.
\newblock URL \url{https://openreview.net/forum?id=yyBis80iUuU}.

\bibitem[Song et~al.(2024{\natexlab{a}})Song, Bagnell, and Singh]{pmlr-v235-song24a}
Song, Y., Bagnell, D., and Singh, A.
\newblock Hybrid reinforcement learning from offline observation alone.
\newblock In Salakhutdinov, R., Kolter, Z., Heller, K., Weller, A., Oliver, N., Scarlett, J., and Berkenkamp, F. (eds.), \emph{International Conference on Machine Learning \CNFX{ICML}}, volume 235 of \emph{Proceedings of Machine Learning Research}, pp.\  46019--46049. PMLR, 21--27 Jul 2024{\natexlab{a}}.
\newblock URL \url{https://proceedings.mlr.press/v235/song24a.html}.

\bibitem[Song et~al.(2024{\natexlab{b}})Song, Swamy, Singh, Bagnell, and Sun]{song2024hypo}
Song, Y., Swamy, G., Singh, A., Bagnell, J., and Sun, W.
\newblock The importance of online data: Understanding preference fine-tuning via coverage.
\newblock In \emph{Advances in Neural Information Processing Systems \CNFX{NeurIPS}}, 2024{\natexlab{b}}.
\newblock URL \url{https://arxiv.org/abs/2406.01462}.

\bibitem[Stiennon et~al.(2020)Stiennon, Ouyang, Wu, Ziegler, Lowe, Voss, Radford, Amodei, and Christiano]{NEURIPS2020_1f89885d}
Stiennon, N., Ouyang, L., Wu, J., Ziegler, D., Lowe, R., Voss, C., Radford, A., Amodei, D., and Christiano, P.~F.
\newblock Learning to summarize with human feedback.
\newblock In Larochelle, H., Ranzato, M., Hadsell, R., Balcan, M., and Lin, H. (eds.), \emph{Advances in Neural Information Processing Systems}, volume~33, pp.\  3008--3021. Curran Associates, Inc., 2020.
\newblock URL \url{https://proceedings.neurips.cc/paper\_files/paper/2020/file/1f89885d556929e98d3ef9b86448f951-Paper.pdf}.

\bibitem[Tajwar et~al.(2024)Tajwar, Singh, Sharma, Rafailov, Schneider, Xie, Ermon, Finn, and Kumar]{tajwar2024preference}
Tajwar, F., Singh, A., Sharma, A., Rafailov, R., Schneider, J., Xie, T., Ermon, S., Finn, C., and Kumar, A.
\newblock Preference fine-tuning of {LLM}s should leverage suboptimal, on-policy data.
\newblock In \emph{Forty-first International Conference on Machine Learning}, 2024.
\newblock URL \url{https://openreview.net/forum?id=bWNPx6t0sF}.

\bibitem[Talmor et~al.(2019)Talmor, Herzig, Lourie, and Berant]{talmor-etal-2019-commonsenseqa}
Talmor, A., Herzig, J., Lourie, N., and Berant, J.
\newblock {C}ommonsense{QA}: A question answering challenge targeting commonsense knowledge.
\newblock In \emph{Conference of the North American Chapter of the Association for Computational Linguistics \CNFX{NAACL}}, pp.\  4149--4158, 2019.
\newblock \doi{10.18653/v1/N19-1421}.
\newblock URL \url{https://aclanthology.org/N19-1421}.

\bibitem[Tan et~al.(2024)Tan, Fan, and Wei]{tan2024hybrid}
Tan, K., Fan, W., and Wei, Y.
\newblock Hybrid reinforcement learning breaks sample size barriers in linear {MDP}s.
\newblock In \emph{Advances in Neural Information Processing Systems \CNFX{NeurIPS}}, 2024.
\newblock URL \url{https://openreview.net/forum?id=bPuYxFBHyI}.

\bibitem[Tang et~al.(2024)Tang, Guo, Zheng, Calandriello, Cao, Tarassov, Munos, Ávila Pires, Valko, Cheng, and Dabney]{tang2024understandingperformancegaponline}
Tang, Y., Guo, D.~Z., Zheng, Z., Calandriello, D., Cao, Y., Tarassov, E., Munos, R., Ávila Pires, B., Valko, M., Cheng, Y., and Dabney, W.
\newblock Understanding the performance gap between online and offline alignment algorithms, 2024.
\newblock URL \url{https://arxiv.org/abs/2405.08448}.

\bibitem[Taori et~al.(2023)Taori, Gulrajani, Zhang, Dubois, Li, Guestrin, Liang, and Hashimoto]{alpaca_stanford}
Taori, R., Gulrajani, I., Zhang, T., Dubois, Y., Li, X., Guestrin, C., Liang, P., and Hashimoto, T.~B.
\newblock Stanford alpaca: An instruction-following llama model.
\newblock \url{https://github.com/tatsu-lab/stanford_alpaca}, 2023.

\bibitem[Team(2023)]{MosaicML2023Introducing}
Team, M.~N.
\newblock Introducing mpt-30b: Raising the bar for open-source foundation models, 2023.
\newblock URL \url{www.mosaicml.com/blog/mpt-30b}.
\newblock Accessed: 2023-06-22.

\bibitem[Touvron et~al.(2023)Touvron, Martin, Stone, Albert, Almahairi, Babaei, Bashlykov, Batra, Bhargava, Bhosale, Bikel, Blecher, Ferrer, Chen, Cucurull, Esiobu, Fernandes, Fu, Fu, Fuller, Gao, Goswami, Goyal, Hartshorn, Hosseini, Hou, Inan, Kardas, Kerkez, Khabsa, Kloumann, Korenev, Koura, Lachaux, Lavril, Lee, Liskovich, Lu, Mao, Martinet, Mihaylov, Mishra, Molybog, Nie, Poulton, Reizenstein, Rungta, Saladi, Schelten, Silva, Smith, Subramanian, Tan, Tang, Taylor, Williams, Kuan, Xu, Yan, Zarov, Zhang, Fan, Kambadur, Narang, Rodriguez, Stojnic, Edunov, and Scialom]{touvron2023llama2}
Touvron, H., Martin, L., Stone, K., Albert, P., Almahairi, A., Babaei, Y., Bashlykov, N., Batra, S., Bhargava, P., Bhosale, S., Bikel, D., Blecher, L., Ferrer, C.~C., Chen, M., Cucurull, G., Esiobu, D., Fernandes, J., Fu, J., Fu, W., Fuller, B., Gao, C., Goswami, V., Goyal, N., Hartshorn, A., Hosseini, S., Hou, R., Inan, H., Kardas, M., Kerkez, V., Khabsa, M., Kloumann, I., Korenev, A., Koura, P.~S., Lachaux, M.-A., Lavril, T., Lee, J., Liskovich, D., Lu, Y., Mao, Y., Martinet, X., Mihaylov, T., Mishra, P., Molybog, I., Nie, Y., Poulton, A., Reizenstein, J., Rungta, R., Saladi, K., Schelten, A., Silva, R., Smith, E.~M., Subramanian, R., Tan, X.~E., Tang, B., Taylor, R., Williams, A., Kuan, J.~X., Xu, P., Yan, Z., Zarov, I., Zhang, Y., Fan, A., Kambadur, M., Narang, S., Rodriguez, A., Stojnic, R., Edunov, S., and Scialom, T.
\newblock {LLAMA 2: Open Foundation and Fine-Tuned Chat Models}.
\newblock \emph{arXiv preprint arXiv:2307.09288}, 2023.
\newblock URL \url{hhttps://arxiv.org/abs/2307.09288}.

\bibitem[Tunstall et~al.(2023)Tunstall, Lambert, Rajani, Beeching, Le~Scao, von Werra, Han, Schmid, and Rush]{Tunstall2023starchat-alpha}
Tunstall, L., Lambert, N., Rajani, N., Beeching, E., Le~Scao, T., von Werra, L., Han, S., Schmid, P., and Rush, A.
\newblock Creating a coding assistant with starcoder.
\newblock \emph{Hugging Face Blog}, 2023.
\newblock https://huggingface.co/blog/starchat.

\bibitem[Wagenmaker \& Pacchiano(2023)Wagenmaker and Pacchiano]{10.5555/3618408.3619878}
Wagenmaker, A. and Pacchiano, A.
\newblock Leveraging offline data in online reinforcement learning.
\newblock In \emph{International Conference on Machine Learning \CNFX{ICML}}, ICML'23. JMLR.org, 2023.

\bibitem[Wang et~al.(2024)Wang, Dong, Delalleau, Zeng, Shen, Egert, Zhang, Sreedhar, and Kuchaiev]{wang2024helpsteer2}
Wang, Z., Dong, Y., Delalleau, O., Zeng, J., Shen, G., Egert, D., Zhang, J.~J., Sreedhar, M.~N., and Kuchaiev, O.
\newblock Helpsteer2: Open-source dataset for training top-performing reward models, 2024.

\bibitem[Wei et~al.(2022)Wei, Wang, Schuurmans, Bosma, Xia, Chi, Le, Zhou, et~al.]{wei2022chain}
Wei, J., Wang, X., Schuurmans, D., Bosma, M., Xia, F., Chi, E., Le, Q.~V., Zhou, D., et~al.
\newblock Chain-of-thought prompting elicits reasoning in large language models.
\newblock \emph{Advances in Neural Information Processing Systems \CNFX{NeurIPS}}, 35:\penalty0 24824--24837, 2022.
\newblock URL \url{https://arxiv.org/abs/2201.11903}.

\bibitem[Wei et~al.(2024)Wei, Liu, and Erichson]{wei2024emojiattackmethodmisleading}
Wei, Z., Liu, Y., and Erichson, N.~B.
\newblock Emoji attack: A method for misleading judge llms in safety risk detection, 2024.
\newblock URL \url{https://arxiv.org/abs/2411.01077}.

\bibitem[Xie et~al.(2024)Xie, Foster, Krishnamurthy, Rosset, Awadallah, and Rakhlin]{xie2024exploratorypreferenceoptimizationharnessing}
Xie, T., Foster, D.~J., Krishnamurthy, A., Rosset, C., Awadallah, A., and Rakhlin, A.
\newblock Exploratory preference optimization: Harnessing implicit q*-approximation for sample-efficient rlhf, 2024.
\newblock URL \url{https://arxiv.org/abs/2405.21046}.

\bibitem[Xiong et~al.(2024)Xiong, Dong, Ye, Wang, Zhong, Ji, Jiang, and Zhang]{xiong2024iterative}
Xiong, W., Dong, H., Ye, C., Wang, Z., Zhong, H., Ji, H., Jiang, N., and Zhang, T.
\newblock Iterative preference learning from human feedback: Bridging theory and practice for rlhf under kl-constraint.
\newblock In \emph{International Conference on Machine Learning \CNFX{ICML}}, 2024.

\bibitem[Xu et~al.(2023)Xu, Sun, Zheng, Geng, Zhao, Feng, Tao, and Jiang]{xu2023wizardlm}
Xu, C., Sun, Q., Zheng, K., Geng, X., Zhao, P., Feng, J., Tao, C., and Jiang, D.
\newblock {WizardLM:} empowering large language models to follow complex instructions.
\newblock \emph{arXiv preprint arXiv:2304.12244}, 2023.
\newblock URL \url{https://arxiv.org/abs/2304.12244}.

\bibitem[Xu et~al.(2024{\natexlab{a}})Xu, Sharaf, Chen, Tan, Shen, Van~Durme, Murray, and Kim]{pmlr-v235-xu24t}
Xu, H., Sharaf, A., Chen, Y., Tan, W., Shen, L., Van~Durme, B., Murray, K., and Kim, Y.~J.
\newblock Contrastive preference optimization: Pushing the boundaries of {LLM} performance in machine translation.
\newblock In Salakhutdinov, R., Kolter, Z., Heller, K., Weller, A., Oliver, N., Scarlett, J., and Berkenkamp, F. (eds.), \emph{Proceedings of the 41st International Conference on Machine Learning}, volume 235 of \emph{Proceedings of Machine Learning Research}, pp.\  55204--55224. PMLR, 21--27 Jul 2024{\natexlab{a}}.

\bibitem[Xu et~al.(2024{\natexlab{b}})Xu, Lee, Sukhbaatar, and Weston]{xu2023things}
Xu, J., Lee, A., Sukhbaatar, S., and Weston, J.
\newblock Some things are more cringe than others: Preference optimization with the pairwise cringe loss.
\newblock In \emph{International Conference on Machine Learning \CNFX{ICML}}, 2024{\natexlab{b}}.

\bibitem[Xu et~al.(2024{\natexlab{c}})Xu, Fu, Gao, Ye, Liu, Mei, Wang, Yu, and Wu]{pmlr-v235-xu24h}
Xu, S., Fu, W., Gao, J., Ye, W., Liu, W., Mei, Z., Wang, G., Yu, C., and Wu, Y.
\newblock Is {DPO} superior to {PPO} for {LLM} alignment? {A} comprehensive study.
\newblock In Salakhutdinov, R., Kolter, Z., Heller, K., Weller, A., Oliver, N., Scarlett, J., and Berkenkamp, F. (eds.), \emph{Proceedings of the 41st International Conference on Machine Learning}, volume 235 of \emph{Proceedings of Machine Learning Research}, pp.\  54983--54998. PMLR, 21--27 Jul 2024{\natexlab{c}}.
\newblock URL \url{https://proceedings.mlr.press/v235/xu24h.html}.

\bibitem[Yuan et~al.(2024)Yuan, Pang, Cho, Li, Sukhbaatar, Xu, and Weston]{yuan2024selfrewarding}
Yuan, W., Pang, R.~Y., Cho, K., Li, X., Sukhbaatar, S., Xu, J., and Weston, J.
\newblock Self-rewarding language models, 2024.
\newblock URL \url{https://arxiv.org/abs/2401.10020}.

\bibitem[Zellers et~al.(2019)Zellers, Holtzman, Bisk, Farhadi, and Choi]{zellers2019hellaswag}
Zellers, R., Holtzman, A., Bisk, Y., Farhadi, A., and Choi, Y.
\newblock {HellaSwag}: {C}an a machine really finish your sentence?
\newblock In \emph{ACL}, 2019.

\bibitem[Zhao et~al.(2024)Zhao, Ren, Hessel, Cardie, Choi, and Deng]{zhao2024wildchat}
Zhao, W., Ren, X., Hessel, J., Cardie, C., Choi, Y., and Deng, Y.
\newblock Wildchat: 1m chat{GPT} interaction logs in the wild.
\newblock In \emph{The Twelfth International Conference on Learning Representations}, 2024.
\newblock URL \url{https://openreview.net/forum?id=Bl8u7ZRlbM}.

\bibitem[Zheng et~al.(2023)Zheng, Chiang, Sheng, Zhuang, Wu, Zhuang, Lin, Li, Li, Xing, et~al.]{zheng2023judging}
Zheng, L., Chiang, W.-L., Sheng, Y., Zhuang, S., Wu, Z., Zhuang, Y., Lin, Z., Li, Z., Li, D., Xing, E., et~al.
\newblock Judging llm-as-a-judge with mt-bench and chatbot arena.
\newblock In \emph{Advances in Neural Information Processing Systems \CNFX{NeurIPS}}, 2023.
\newblock URL \url{https://arxiv.org/abs/2306.05685}.

\bibitem[Zhou et~al.(2023)Zhou, Lu, Mishra, Brahma, Basu, Luan, Zhou, and Hou]{zhou2023instruction}
Zhou, J., Lu, T., Mishra, S., Brahma, S., Basu, S., Luan, Y., Zhou, D., and Hou, L.
\newblock Instruction-following evaluation for large language models.
\newblock \emph{arXiv preprint arXiv:2311.07911}, 2023.

\end{thebibliography}
\bibliographystyle{icml2025}


\newpage
\onecolumn
\appendix

\begin{center}
{\Large \textbf{Supplemental Material}}
\end{center}

\section{Hyperparameters}
\label{appendix:hparams}

For all of our experiments, we use OpenRLHF \cite{hu2024openrlhf} for training and lm-evaluation-harness \cite{eval-harness} for evaluation. Initially, we performed hyperparameter sweeps for $\beta = \{0.01, 0.5, 0.1, 1, 5\}$ and $\text{max learning rate} = \{5e-7, 1e-6, 5e-6\}$ for initial exploration for DPO. We report the result on $\beta = 0.1$ and $\text{max learning rate} = 5e-7$ for \textbf{all} of our experiments. 

\section{Alpaca Eval 2.0 Prompt Categories}
\label{appendix:alpaca_eval}
\begin{figure}[h!]
    \vspace{-2pt}
    \centering
    \includegraphics[scale=0.5]{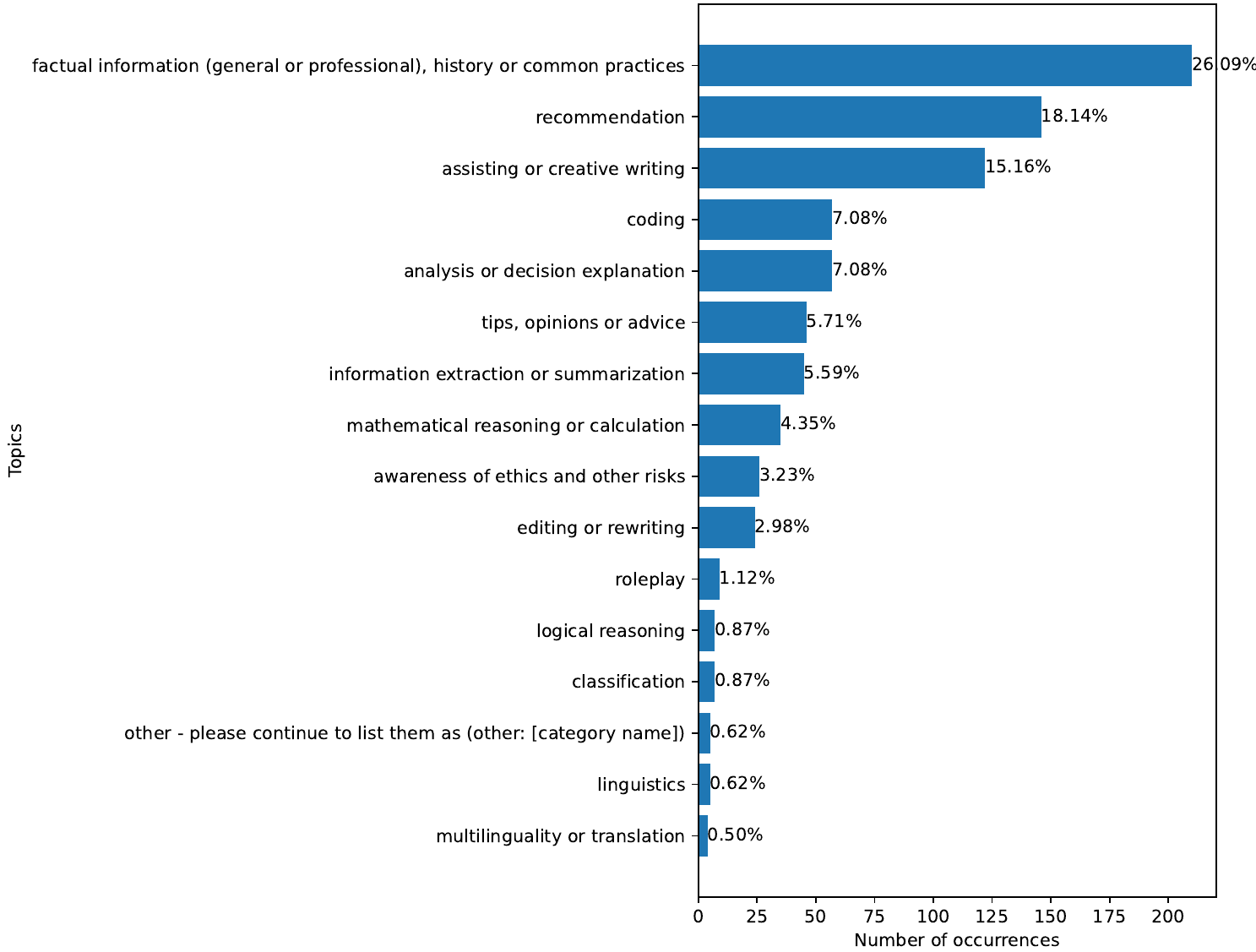}
    \caption{Distribution of different categories of prompts in Alpaca Eval 2.0 \cite{dubois2023alpacafarm, dubois2024lengthcontrolled}.}
    \label{fig:alpaca_prompt_categories}
\end{figure}

We define the category of prompts as the same as WildChat \cite{zhao2024wildchat}, which contains in total 16 categories. We prompt \texttt{gpt-4o-mini} for classifying prompts into one of the 16 categories. We report the distribution of 805 prompts in Alpaca Eval 2.0 \cite{dubois2023alpacafarm, dubois2024lengthcontrolled} in Figure \ref{fig:alpaca_prompt_categories}.

\section{Dataset Details}
\label{appendix:dataset_details}

Here we describe the datasets we used in \S \ref{Section:4}: Ultrafeedback \cite{pmlr-v235-cui24f} and HelpSteer2 \cite{wang2024helpsteer2} in detail:

\begin{itemize}
    \item \textbf{Ultrafeedback \cite{pmlr-v235-cui24f}}: \textbf{Prompts} included questions in TruthfulQA \cite{lin2021truthfulqa}, FalseQA \cite{hu-etal-2023-wont}, Evol-Instruct \cite{xu2023wizardlm}, UltraChat \cite{ding2023enhancing}, ShareGPT \cite{chiangchatbot}, and the FLAN \cite{longpre2023flan} collection; \textbf{Responses} are sampled from a model pool with 17 models: GPT-4, gpt-3.5-turbo, Bard, UltraLM-13B/65B \cite{ding2023enhancing}, WizardLM-7Bv1.1/13B-v1.2/70B-v1.1 \cite{xu2023wizardlm}, Vicuna-33B-v1.3 \cite{ding2023enhancing}, LLaMA2-7B/13B/70B-Chat \cite{touvron2023llama2}, Alpaca-7B \cite{alpaca_stanford}, MPT-30B-Chat \cite{MosaicML2023Introducing}, Falcon-40B-Instruct \cite{falcon40b}, StarChat \cite{Tunstall2023starchat-alpha}, and Pythia-12B \cite{biderman2023pythia}. 
    \item \textbf{HelpSteer2 \cite{wang2024helpsteer2}}: \textbf{Prompts} mainly (over 95\%) are sampled from ShareGPT \cite{chiangchatbot}, with a small portion from proprietary user prompts; \textbf{Responses} are sampled from Nemotron-2/3/4 \cite{nvidia2024nemotron4340btechnicalreport},  Mixtral-8x7B-Instruct-v0.1 \cite{Jiang2023Mistral7}, and human annotators.
\end{itemize}

\section{Evaluation Details}
\label{appendix:eval_details}
We use the \texttt{lm-eval-harness} \cite{eval-harness}\footnote{https://github.com/EleutherAI/lm-evaluation-harness} library for evaluation on all 8 benchmarks (ARC-Challenge, Commonsense QA, Hellaswag, MMLU, OpenQA, PiQA, Winograde, and Ifeval).

The details of the 8 benchmarks are listed below:

\begin{itemize}
    \item \textbf{ARC-Challenge \cite{clark2018think}:} A set of multiple-choice questions focusing on reasoning and scientific knowledge across a range of domains.
    \item \textbf{Commonsense QA \cite{talmor-etal-2019-commonsenseqa}:} A benchmark designed to evaluate a model's ability to answer questions based on commonsense knowledge, requiring reasoning beyond factual recall.
    \item \textbf{Hellaswag \cite{zellers2019hellaswag}:} A benchmark focused on commonsense reasoning and narrative understanding, where models predict the most plausible continuation of an incomplete story.
    \item \textbf{MMLU \cite{hendrycks2020measuring}:} The Massive Multitask Language Understanding benchmark evaluates models on a diverse set of tasks, including STEM, humanities, and social sciences, measuring performance on human-level tasks.
    \item \textbf{OpenQA \cite{mihaylov2018can}:} A benchmark for evaluating models on open-domain question answering, testing how well they retrieve and reason over information to answer diverse queries.
    \item \textbf{PiQA \cite{bisk2020piqa}:} A benchmark focused on physical commonsense reasoning, requiring models to predict the most plausible solution to problems involving physical world interactions.
    \item \textbf{Winograde \cite{sakaguchi2020winogrande}:} A benchmark that assesses models' ability to perform common-sense reasoning in co-reference tasks, where ambiguous pronouns are resolved using contextual clues.
    \item \textbf{Ifeval \cite{zhou2023instruction}:} A benchmark designed to assess models on their ability to follow precise instructions that are verifiable.
\end{itemize}
In Alpaca Eval 2.0, for reproducibility, we use greedy decoding with a max length of 2048 tokens. 

\section{Full Results}
\label{appendix:full_results}
Table \ref{tab:llama3_full_results} and \ref{tab:tulu_full_results} shows the full results on all 8 benchmarks: Commonsense QA (CQA; \citep{talmor-etal-2019-commonsenseqa}, Hellaswag (HS; \cite{zellers2019hellaswag}, Openbook QA (OQA; \citep{OpenBookQA2018}, PIQA \citep{bisk2020piqa}, Winograde \citep{sakaguchi2020winogrande}, ARC-Challenge (ARC-C; \citep{clark2018think}), MMLU \citep{hendrycks2020measuring}, and Ifeval \cite{zhou2023instruction}, respectively.

\begin{table}[h!]
\centering
\begin{tabular}{@{}lccccccccc@{}}
\toprule
                      & ARC-C & CQA   & HSwag  & MMLU  & OQA   & PiQA  & WG & Ifeval   & Avg. Score \\ \midrule
\texttt{Llama-3.1-8B-Instruct} & 51.90 & 77.14 & 59.14 & 67.93  & 33.40 & 79.97 & 73.55 & 44.00 & 60.88      \\ \midrule
 \rowcolor[gray]{0.9}     \multicolumn{10}{c}{ \textit{Experiments on UltraFeedback \cite{pmlr-v235-cui24f}}}  \\ \midrule
Off-policy DPO        & 53.07 & 77.40 & 59.33 & 68.15  & 34.40 & 80.47 & 74.11 & 46.40 & 61.67      \\
On-policy DPO         & 54.18 & 77.81 & 59.88 & 68.60  & 34.80 & 80.41 & 74.98 & 46.03 & 62.09      \\
HyPO \cite{song2024hypo}                 & 54.86 & 77.64 & 60.09 & 68.20  & \textbf{35.40} & 80.63 & 74.59 & 44.60 & 62.00      \\
DPO-Mix-P \cite{shi2024crucialrolesamplersonline}           & 53.07 & \textbf{77.89} & 59.67 & 68.37  & 34.00 & 80.20 & 74.03 & 45.66 & 61.61      \\
\rowcolor{cyan!20} \method     & \textbf{55.03} & 77.81 & \textbf{61.08} & \textbf{69.47}  & \textbf{35.40} & \textbf{81.69} & \textbf{75.00} & \textbf{48.98} & \textbf{63.06}      \\ \midrule
 \rowcolor[gray]{0.9}     \multicolumn{10}{c}{ \textit{Experiments on HelpSteer2 \cite{wang2024helpsteer2}}}  \\ \midrule
 Off-policy DPO        & 51.79 & 76.99 & 59.22 & 68.11  & 33.60 & 80.09 & 73.88 & \textbf{48.61} & 61.54      \\
On-policy DPO         & 53.58 & 77.48 & 59.63 & \textbf{68.45}  & 35.00 & 80.47 & 74.11 & 45.47 & 61.77      \\
HyPO                  & 53.50 & 76.82 & 59.37 & 68.32  & 34.00 & 80.30 & 73.88 & 47.32 & 61.69      \\
DPO-Mix-P             & 53.49 & 77.31 & 59.55 & 68.38  & \textbf{34.60} & 79.86 & 74.34 & 44.36 & 61.49      \\
\rowcolor{cyan!20} \method & \textbf{54.27} & \textbf{78.05} & \textbf{59.76} & 68.39  & \textbf{34.60} & \textbf{80.20} & \textbf{74.51} & 47.21 & \textbf{62.12}   \\
 \bottomrule
\end{tabular}
\caption{Detailed Benchmark results (ARC Challenge, Common QA, Hellaswag, MMLU, Openbook QA, PiQA, Winograde, Ifeval) on training \texttt{Llama-3.1-8B-Instruct} \cite{dubey2024llama3herdmodels} on the Ultrafeedback \cite{pmlr-v235-cui24f} and HelpSteer2 \cite{wang2024helpsteer2} dataset.}
\label{tab:llama3_full_results}
\end{table}

\begin{table*}[h!]
\centering
\begin{tabular}{@{}lccccccccc@{}}
\toprule
Model & ARC-C & CQA & HSwag & MMLU & OQA & PiQA & WG & Ifeval & Avg. Score \\
\midrule
\texttt{Llama-3.1-Tulu-3-8B-SFT} & 52.73 & 75.76 & 61.87 & 63.67 & 36.80 & 80.79 & 74.35  & 31.79 & 59.72 \\ \midrule
 \rowcolor[gray]{0.9}     \multicolumn{10}{c}{ \textit{Experiments on UltraFeedback \cite{pmlr-v235-cui24f}}}  \\ \midrule
Off-policy DPO        & 55.20                & 76.49                & 62.94                & 58.77                & 37.60                & 80.96                & 74.19                & 33.97                & 60.02                \\
On-policy DPO         & 54.18                & 76.17                & 62.41                & 58.70                 & 37.60                & 80.85                & \textbf{74.27}                & 33.40                & 59.70                \\
HyPO \cite{song2024hypo}                 & 55.15                & 76.33                & 61.91                & 62.17                & 37.40                & 80.19                & 73.95                & 33.10                & 60.03                \\
DPO-Mix-P \cite{shi2024crucialrolesamplersonline}            & \textbf{56.32}                & 76.02                & 63.01                & \textbf{64.21}                & \textbf{39.50}                & 80.36                & 73.11                & 33.10                & 60.70                \\
\rowcolor{cyan!20} \method       & 56.06                & \textbf{76.82}                & \textbf{63.13}                & 63.60                 & 38.60                & \textbf{81.23}                & 74.03                & \textbf{35.62}                & \textbf{61.14}                \\ \midrule
 \rowcolor[gray]{0.9}     \multicolumn{10}{c}{ \textit{Experiments on HelpSteer2 \cite{wang2024helpsteer2}}}  \\ \midrule 
Off-policy DPO        & 52.90                 & 76.09                & 61.83                               & 63.68                & 37.00                   & \textbf{80.79}                & 73.80            & 33.46      & 59.94            \\
On-policy DPO & 51.90                 & 75.92                & 61.88                             & 63.77                & 37.00                   & 80.74                & 74.03           & 36.41     & 60.21                \\
HyPO \cite{song2024hypo} & 52.65                & \textbf{76.33}                & 61.85                              & \textbf{63.90}                 & 36.40                 & 80.63                & \textbf{74.43}          & 33.46       & 59.96             \\ 
DPO-Mix-P \cite{shi2024crucialrolesamplersonline} & 52.55 & 76.26 & 62.11    & 63.71 & 36.20  & 80.19 & 74.17 & 35.10 & 60.04  \\
\rowcolor{cyan!20} \method  & \textbf{52.90}                 & 76.09                & 63.84                                & 63.84               & \textbf{37.40}                 & 80.74                & 74.11        & \textbf{40.85}        & \textbf{61.22}  \\         
\bottomrule
\end{tabular}
\caption{Detailed Benchmark results (ARC Challenge, Common QA, Hellaswag, MMLU, Openbook QA, PiQA, Winograde, Ifeval) on training \texttt{Llama-3.1-Tulu-3-8B-SFT} \cite{lambert2024tulu3pushingfrontiers} on the Ultrafeedback \cite{pmlr-v235-cui24f} and HelpSteer2 \cite{wang2024helpsteer2}.}
\label{tab:tulu_full_results}
\end{table*}

\section{Additional Related Works}
\label{appendix:additional_related_works}

\paragraph{Language Model Alignment} 
Alignment of language models is typically done at the post-training stage, in order to make the language model prefer certain types of responses. In our paper, we specifically refer the stage \textbf{after} supervised fine-tuning (SFT) as to ``alignment". In this stage, two paradigms are typically employed: In Reinforcement Learning with Human Feedback (RLHF; \cite{NEURIPS2020_1f89885d, ouyang2022training}), an on-policy alignment method assigns a reward to on-policy rollouts. Such a reward can be of a separate reward model trained on off-policy generations \cite{ouyang2022training}, hand-crafted heuristics \cite{bai2022constitutional}, or even AI feedback \cite{guo2024directlanguagemodelalignment}. The other paradigm does not require on-policy rollouts and directly manipulates the log probabilities of the offline chosen and rejected responses, hoping that the model generalizes from these offline responses to learn what principle it should follow. Direct Preference Optimization (DPO; \cite{rafailov2024direct}) is the most canonical one that aims to push the log-probs of chosen responses higher and the log-probs of rejected responses lower. Many variants of DPO have been proposed \cite{pal2024smaugfixingfailuremodes, meng2024simpo, pmlr-v238-gheshlaghi-azar24a, pmlr-v235-xu24t} but they mostly used \textbf{off-policy} examples. However, the contrastive nature of the DPO loss can also take on-policy examples. Therefore, many on-policy variants of DPO have also been proposed \cite{xu2023things, yuan2024selfrewarding, xiong2024iterative, guo2024directlanguagemodelalignment}. Furthermore, there are also off-policy works that are non-contrastive \cite{ethayarajhmodel}. Our paper directly studies the difference (\S \ref{Section:3}) and interaction (\S \ref{Section:4}) of on- vs. off-policy data, disentangling the effect of the contrastive nature of DPO and the non-contrastive nature of standard RLHF. 

\section{Per-task Performance of \method}
\label{appendix:sm_per_task}
\begin{figure*}[h!]
\centering
    \begin{subfigure}[b]{0.44\textwidth}
    \centering
    \includegraphics[scale=0.45]{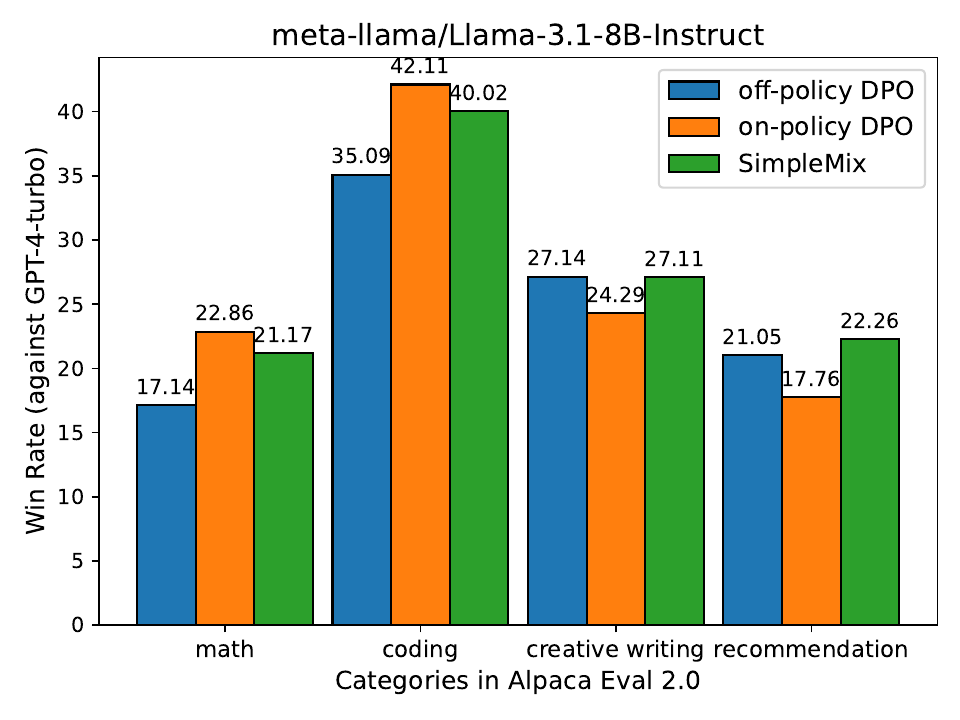}
    \label{fig:llama3_per_category_sm}
    \end{subfigure}
    \hspace{0.02\textwidth}
    \begin{subfigure}[b]{0.44\textwidth}
    \centering
    \includegraphics[scale=0.45]{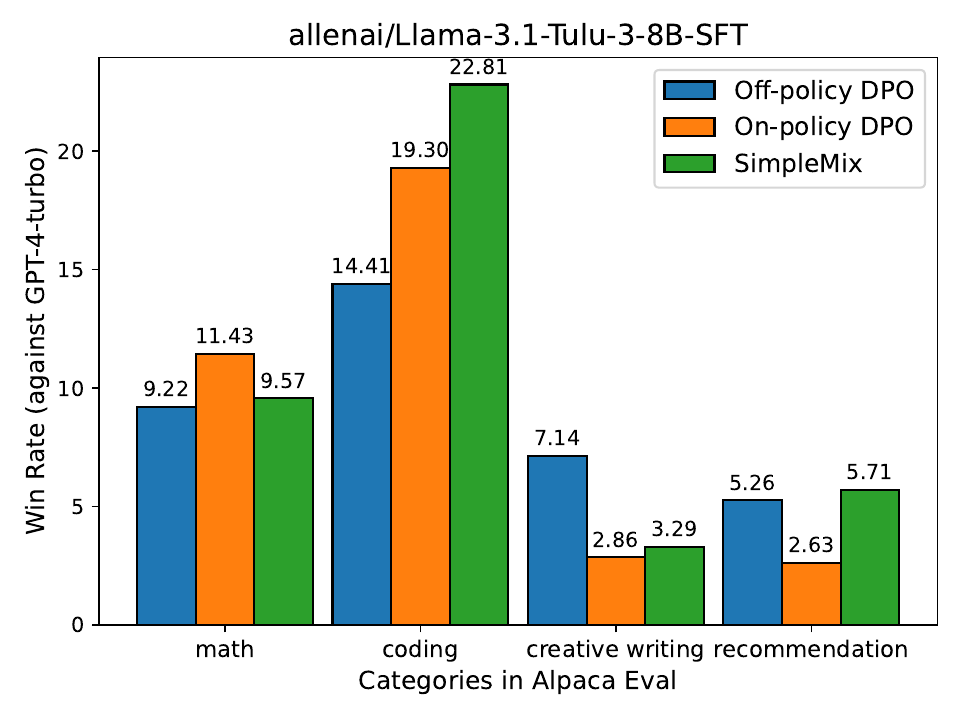}
    \label{fig:tulu_per_category_sm}
    \end{subfigure}
    \caption{Comparison of win rates (against GPT-4-turbo) across different prompt categories in Alpaca Eval 2.0 by training \texttt{Llama-3.1-8B-Instruct} and \texttt{Tulu-3-8B-SFT} with \method. \method~offers a good balance between reasoning tasks and open-ended tasks.}
    \label{fig:winrates_by_category_with_sm}
\end{figure*}

\newpage 
\section{Diverse Prompt}
\label{appendix:diverse_prompt}

Figure 8 shows the prompt we used to elicit the language model to generate a response different from previous responses.

\begin{figure*}[h]
\begin{tcolorbox}[breakable,colback=yellow!6!white,colframe=blue!50!green]
Please generate another response that is different from all the previous ones. \\
It can differ in aspects such as but not limited to tone, formality, strategy, judgment, conciseness, structure. It can also involve asking for clarity if the input prompt is unclear.
\end{tcolorbox}
\label{fig:diverse_prompt}
\caption{The prompt used to elicit diverse responses from a language model.}
\end{figure*}

\section{Full Reward Distributions}

\begin{figure}
    \centering
    \includegraphics[scale=0.5]{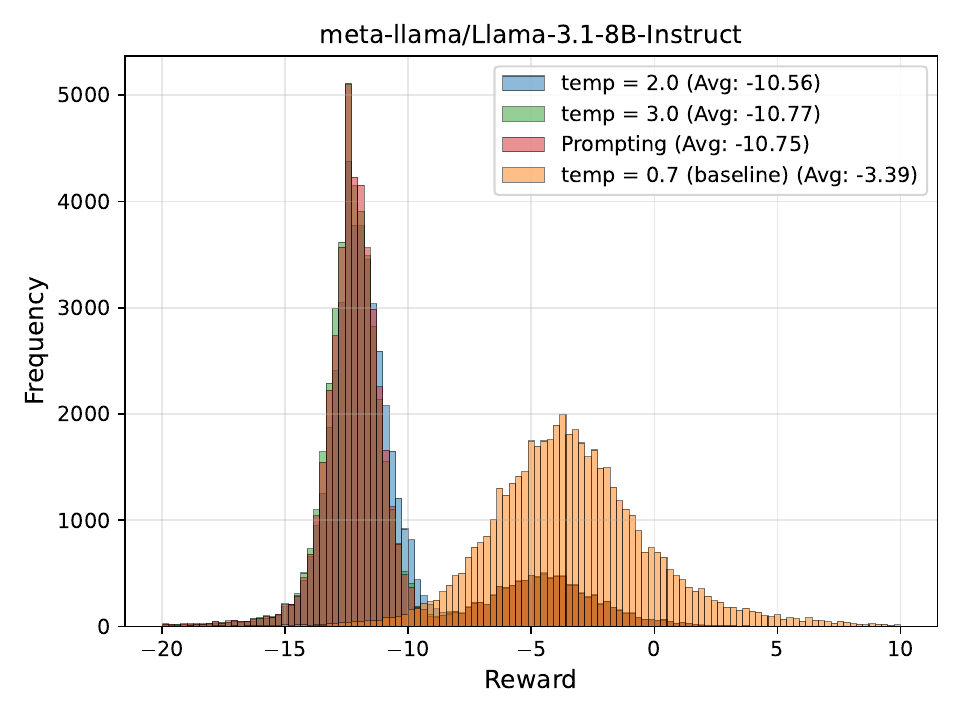}
    \caption{Histogram of reward distribution for different sampling temperatures. Increasing the temperature results in generations of lower quality, as measured by our reward model.}
    \label{fig:reward_histogram}
\end{figure}

\begin{figure}
    \centering
    \includegraphics[width=\linewidth]{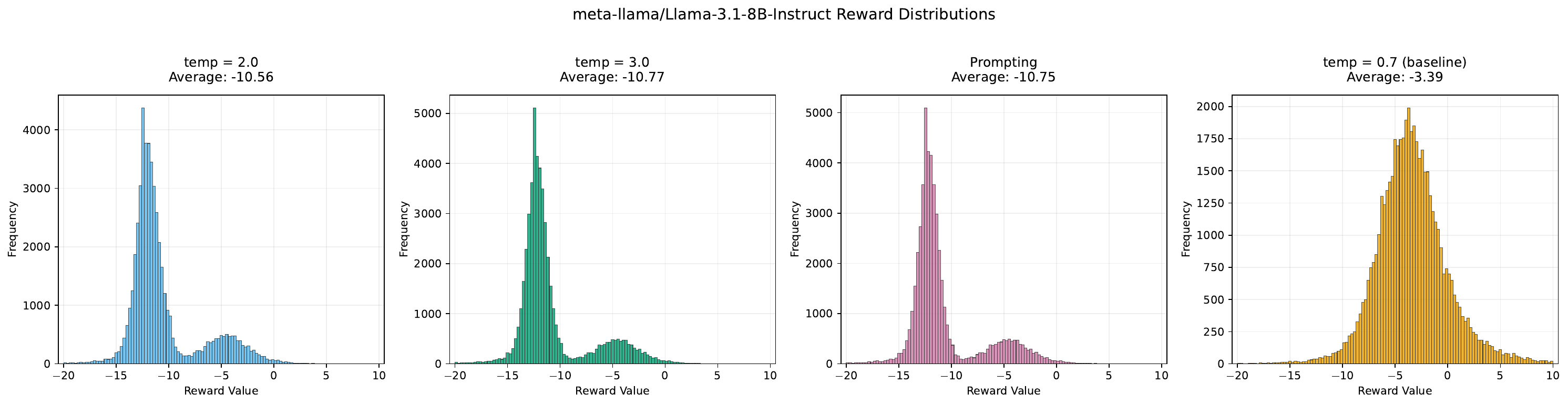}
    \caption{Histograms of the reward for sampling from \texttt{Llama-3.1-8B-Instruct} with different sampling schemes (prompting \& and increasing the temperature).}
    \label{fig:histogram_reward}
\end{figure}

Figure \ref{fig:reward_histogram} shows the aggregated histogram of reward of the responses sampled from \texttt{Llama-3.1-8B-Instruct} with sampling temperature $\tau = \{0.7,2,3\}$ and prompting it to generate diverse responses (Prompting). Figure \ref{fig:histogram_reward} shows the individual histograms.

\label{appendix:full_histogram}

\end{document}